\pdfoutput=1 

\documentclass[a4paper,fleqn]{cas-dc}
\usepackage{hyperref}
\usepackage{cleveref}
\usepackage[numbers,sort&compress]{natbib}

\usepackage{colortbl} 
\definecolor{mycolor}{RGB}{235, 235, 235} 
\definecolor{baise}{RGB}{255,255,255}
\usepackage{multirow}
\usepackage{graphicx}
\usepackage{float}
\usepackage{subfig} 
\usepackage{caption}
\usepackage{algpseudocode}
\usepackage{booktabs} 
\usepackage{threeparttable} 
\usepackage{tabularx} 
\usepackage{pifont}
\usepackage{algorithm}
\usepackage{xurl} 

\def\tsc#1{\csdef{#1}{\textsc{\lowercase{#1}}\xspace}}
\tsc{WGM}
\tsc{QE}
\newcommand{\Deltahigh}{\Delta_{\mathrm{high}}}
\newcommand{\E}{\mathbb{E}}
\newcommand{\Prob}{\mathbb{P}}

\newcommand{\Gcal}{\mathcal{G}}

\newtheorem{theorem}{Theorem}
\newtheorem{lemma}{Lemma}
\newtheorem{corollary}{Corollary}
\newtheorem{assumption}{Assumption}
\newtheorem{proposition}{Proposition}

\newtheorem{remark}{Remark}

\makeatletter
\def\@opargbegintheorem#1#2#3{\trivlist
  \let\baselinestretch\@blstr
      \item[\hskip \labelsep{\bfseries #1\ #2}]{\bfseries\upshape (#3).}\itshape}
\makeatother

\begin{document}
\let\WriteBookmarks\relax
\def\floatpagefraction{1}
\def\textfraction{.001}      
\let\printorcid\relax 

\makeatletter
\AtEndDocument{\clearpage\immediate\write\@auxout{\string\csxdef{lastpage}{\number\numexpr\value{page}-1\relax}}}
\makeatother

\shorttitle{BARS: Budget-Aware Sequential Sampling with Refusal}    

\shortauthors{Shenxi Liu et al.}

\title[mode=title]{BARS: Budget-Aware Sequential Sampling with Refusal for Majority Identification under Unknown Distributions}

\author[1]{Shenxi Liu}
\author[1]{Yiwei Li}
\author[2]{Mingyang Zhao}
\author[1]{Yuhang Tian}
\author[1]{Kan Li}
\cormark[1]
\ead{likan@bit.edu.cn}

\address[1]{Beijing Institute of Technology, Beijing, 100081, China} 
\address[2]{Department of Computing, The Hong Kong Polytechnic University, 999077, Hong Kong} 
\cortext[1]{Corresponding author} 

\begin{abstract}
Repeated sampling with majority selection is an effective means of enforcing self-consistency in large language models.
Recently, a variant has also been introduced that terminates early to save costs when the output distribution is favorable.
But it still cannot yield a correct confidence estimate from the answer distribution.
For this issue, we formalize this as majority identification under an unknown distribution with a budget and a target error rate, and propose BARS, a sequential rule certifying the top-two gap and returning an early stop, a budgeted majority vote, or a refusal when feasibility cannot be established. 
A refusal is a statement about the rule and its budget, not a claim that a question is unanswerable, and it tells practitioners when to widen the budget or switch models. 
A certified mode carries proven bounds on the joint answering error and on refusing an instance the budget could handle, while a practical mode, used in our experiments, replaces them with empirical calibration. 
On synthetic instances drawn from known distributions over thousands of seeds, the difficulty-driven behavior is recovered and the price of the certified mode is measured.
On GSM8K the practical mode draws 22.3\% fewer samples than fixed-8 sampling over the same pool, refused questions included, at the cost of declining 18.8\% of the questions.  
On MATH, it greatly improves the accuracy of its output answers into 82.5\% at the cost of honestly refusing to answer half of the difficult questions that the model itself is unable to solve.
\end{abstract}



\maketitle

\section{Introduction}\label{sec:intro}


Large language models (LLMs) are increasingly embedded in expert and decision-support systems. 
In such systems a single answer is often produced by several stochastic decodes of the same prompt, and the majority of those decodes is taken as the consensus, a procedure known as self-consistency \citep{wang2023selfconsistency}. 
Self-consistency markedly improves reasoning accuracy on benchmarks such as GSM8K \citep{cobbe2021} and MATH \citep{hendrycks2021}, yet it applies the same number of decodes to every question. 
This paper addresses the deployment question that follows: \emph{how many samples are enough to trust an answer, and when should the system refuse to answer at all?} 

\begin{figure*}[t]
  \centering
  \begin{minipage}[b]{0.46\textwidth}
    \centering
    \includegraphics[width=\textwidth]{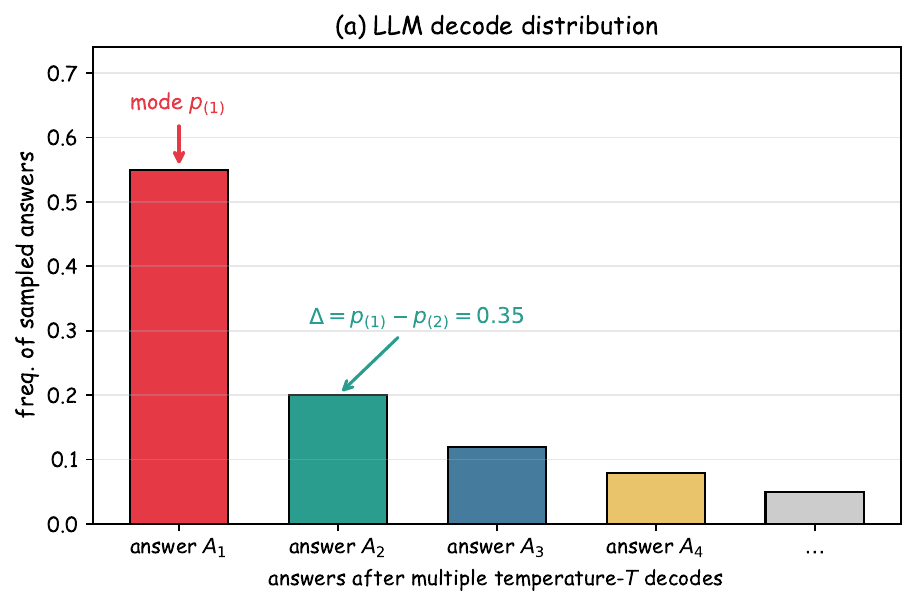}
  \end{minipage}\hfill
  \begin{minipage}[b]{0.50\textwidth}
    \centering
    \includegraphics[width=\textwidth]{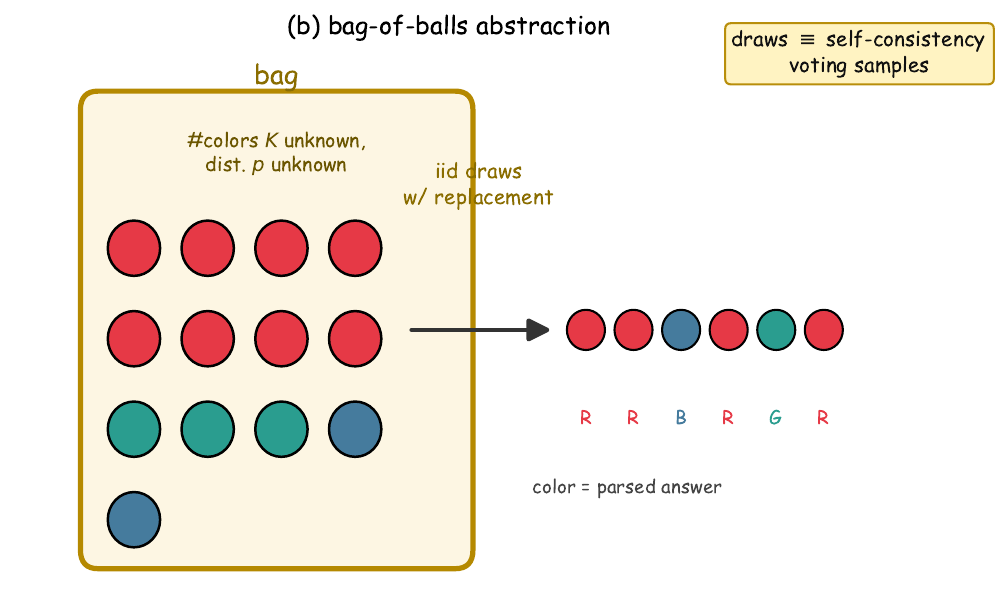}
  \end{minipage}
  \caption{The bag-of-balls abstraction. (a)~Answer distribution obtained by repeatedly decoding the same question at temperature $T$; the top-two gap $\Delta=p_{(1)}-p_{(2)}$ measures the separation between the two most probable answers. (b)~Each decode is viewed as drawing a ball, with replacement, from a bag whose size and composition are unknown; the color of the ball is the parsed answer.}
  \label{fig:bag}
\end{figure*}

A fixed number of decodes fails in two opposite directions, and both failures are common in deployment. 
When the answer distribution is concentrated on one answer, a few samples already give the accuracy obtained with the full budget, so the remaining samples produce no additional evidence. 
When the distribution is spread over several answers, further samples still separate them only slowly, and the majority vote becomes a guess. 
A budget-aware system should therefore adapt its sampling strategy to each question and declare when a reliable answer cannot be produced within the budget. 

We refer to the two decisions as \emph{the stopping decision} (whether the current evidence is already sufficient) and \emph{the refusal decision} (whether the budget can ever be sufficient for this question). 
Section~\ref{sec:method} formalizes both, and the remainder of the paper is organized so that each result can be traced to the decision it supports. 


We reduce the deployment problem to a statistical primitive with an unknown answer-space size and unknown answer probabilities: \emph{drawing balls from a bag with replacement, and identifying the most probable color}. 
We call this the \emph{bag-of-balls abstraction} (Figure~\ref{fig:bag}): each decode draws one ball, and a color is the parsed answer. 
The abstraction assumes neither the number of colors nor the shape of the distribution, and two consequences follow. 
First, it applies to any question whose decodes can be mapped to a finite candidate set and to any model that can be sampled repeatedly. 
Second, classical probability tools cover the rest of the analysis, because the difficulty of the problem is captured by a single quantity: the top-two gap $\Delta = p_{(1)} - p_{(2)}$. 

Formally, we study the following problem. 
\begin{quote}
Within a budget of at most $E^*$ samples, identify the majority of an unknown distribution with error probability at most $p^*$, conditional on answering; when this target cannot be met, the system must report that it is infeasible.
\end{quote}
The analysis assumes independent and identically distributed samples and a unique most probable answer ($\Delta>0$). 
It does not prescribe a distributional shape, but two further conditions bind at the boundaries of the results and are stated where they first matter: the certified width mode requires a known upper bound $\bar K$ on the number of colors (Assumption~\ref{asm:color-bound}), and every guarantee concerns the \emph{majority} answer rather than the correct one. 
The second condition deserves emphasis: if the model's most probable response is wrong, the rule identifies it with the same confidence it would apply to a correct answer, and nothing in the analysis detects the difference. 


We propose \textbf{BARS} (\textbf{B}udget-\textbf{A}ware \textbf{R}efusal \textbf{S}ampling), a sequential stopping rule that maintains the empirical top-two gap $\hat\Delta_t$ and its plug-in variance $\hat\sigma_t^2$ and certifies distributional difficulty with time-uniform confidence sequences \citep{howard2021,waudbysmith2023,darling1967}. 
The rule returns one of three outputs. 
The \emph{fast branch} stops after $w$ consecutive same-color samples, where the run length $w$ is the smallest admissible value for which the worst-case error rate is at most $p^*$, computed by a Markov dynamic program. 
The \emph{naive branch} spends the remaining budget and takes the majority, with a Hoeffding bound controlling its error. 
The \emph{refusal branch} outputs ``infeasible within budget'' when the rule's own criterion cannot establish feasibility; its rate of firing on instances the budget could have handled is bounded by $\delta$ plus an explicit, empirically calibrated plug-in bias term. 
We treat refusal as an informative third output rather than a failure, because it lets a deployed system escalate to a human, widen the budget or switch the underlying model. 

Two width modes parameterize the same rule and are kept apart throughout the paper. 
The \emph{certified mode} uses a time-uniform half-width whose constants are explicit, so that the trigger and the resulting error bounds hold with probability at least $1-\delta$; it is the mode covered by the theorems. 
The \emph{practical mode} uses a variance-scaled half-width with a fixed multiplier that controls the deviation only at the current sample size; it carries no theorem-level guarantee and is calibrated empirically, and it is the mode used in our language-model experiments. 
Each of the three outputs is therefore tied to a declared configuration, and every claim in this paper is marked with the configuration under which it applies. 


Our contributions are as follows. 
\begin{enumerate}
  \item \textbf{A budget-aware stopping and refusal rule with explicit configuration.} 
  BARS drives its sampling cost by the certified difficulty of the instance and returns either an answer or a refusal, and it separates two width modes: the certified mode, for which the mis-trigger probability over the whole exploration process is bounded by $\delta$, and the practical mode, which trades that bound for a higher answer rate. 
  \item \textbf{Guarantees on answering and on refusing, together with the boundary of the analysis.} 
  For the certified mode with the pessimistic chooser, we prove a bound of $p^*+\delta$ on the joint event that the rule answers and answers wrongly, a bound of $\delta + B_{\mathrm{plug}}$ on the joint event that it refuses an instance the budget could have handled, an exact decomposition of the expected number of samples, and a lower bound showing that no rule can certify the majority of the two-point family with fewer than $c\,\Delta^{-2}\ln(1/\delta)$ samples in expectation. 
  We do not claim a matching upper bound, and we prove the lower bound only for the two-point family ($K=2$). 
  \item \textbf{An empirical account of when the budget buys what.} 
  {\setlength{\emergencystretch}{1.5em}%
  Synthetic instances over thousands of seeds show the difficulty-driven behavior predicted by the certified-mode analysis, and real-LLM runs on GSM8K and MATH (Qwen2.5-7B, a pre-generated pool of sixteen decodes per question) show the same qualitative pattern at a deployment-scale budget, where the certified mode is provably unreachable and the practical mode is what a practitioner would actually run.
  A discussion of those runs separates what the two settings establish from what they do not.\par} 
\end{enumerate}

What we do not claim is as important as what we do. 
Our guarantees concern the joint events of answering wrongly and of refusing a handleable instance, not the conditional error rate on answered questions alone, which is smaller only when the answer rate is close to one. 
The certified bounds hold for the certified width with the pessimistic chooser; the practical mode is calibrated empirically, and $B_{\mathrm{plug}}$ has no distribution-free upper bound, because none exists for a plug-in criterion of this kind. 
Finally, a refusal is a statement about this rule and this budget, not a proof that the question is unanswerable. 

The remainder of the paper is organized as follows. 
Section~\ref{sec:related} compares BARS with the three lines of work it sits between. 
Section~\ref{sec:method} formalizes the decision problem, states the two width modes and gives the rule in pseudocode. 
Section~\ref{sec:theory} states the guarantees, in the order in which they depend on one another. 
Section~\ref{sec:experiments} reports the experiments, in two settings that share one design and one competitor set, and then states what the two establish together and what neither establishes.
Section~\ref{sec:conclusion} concludes.

\section{Related work}\label{sec:related}

BARS sits between three lines of work: adaptive test-time sampling, selective prediction with abstention, and best-arm identification with abstention. 
We compare BARS with each line of work and close with a table that separates them along five dimensions, so that the boundary of this paper can be read off directly rather than assembled from the prose. 
The analysis itself builds on standard tools --- Hoeffding's inequality \citep{hoeffding1963}, the empirical Bernstein bound \citep{maurer2009}, time-uniform confidence sequences \citep{howard2021,waudbysmith2023,darling1967} and the classical theory of runs \citep{feller1968} --- and we do not present any of them as a new idea. 
What the paper contributes is how they are assembled into a decision rule with an explicit refusal output, and what that assembly costs. 

\subsection{Test-time scaling and adaptive sampling}\label{sec:rel-tts}

Self-consistency \citep{wang2023selfconsistency} is the fixed-budget baseline for LLM majority voting, and its error analysis typically assumes a finite, known answer space \citep{wang2023selfconsistency}. 
Adaptive variants include early-stopping self-consistency (ESC) \citep{esc2024}, difficulty-adaptive self-consistency \citep{difficulty2024}, CGES \citep{cges2025} and sequential-evidence DASE \citep{dase2026}. 
These methods stop on heuristics that yield empirical savings but carry no provable control of the error rate or of the mis-trigger rate, and none of them treats refusal as an output. 
Two further groups are close in method but not in objective: test-time scaling studies how accuracy grows with the number of samples when the answer space is known exactly \citep{llmmonkeys2024,snell2024,chen2021passk}, and pass-$k$ analyses score the sample set rather than selecting a single answer \citep{chen2021passk,passkidentify2026}. 
The prior work closest to ours in problem setting is MARS \citep{mars2026}, which stops on parallel test-time scaling checkpoints using a margin-adversarial criterion. 
MARS differs in three respects that matter for deployment: its guarantee is heuristic rather than derived under a stated confidence level; it has no ``infeasible within budget'' output; and it targets parallel traces, whereas we replay the pool as a stream and therefore allow a single sampler. 

\subsection{Selective prediction, abstention, and agreement-based selection}\label{sec:rel-conformal}

Selective prediction lets a model abstain when it is uncertain. 
The SelectiveNet family \citep{selectivenet2019,abstainunreliable2022} trades coverage against risk through a threshold or a training objective, and conformal abstention \citep{conformalabstention2024} with its successors, such as SConU \citep{sconu2025} and conformal risk control (CRC) \citep{crc2026}, controls the risk of the abstention decision itself. 
In all of these methods the input is a batch of responses that has already been drawn, and the decision is a single use-or-not choice; there is no sequential sampling decision, and hence no notion of a budget that the decision layer itself consumes. 
CRC makes this boundary sharp: its impossibility result \citep{crc2026} shows that whenever the base risk exceeds a threshold, any distribution-free method under a coverage constraint must abstain on a fixed fraction of the inputs. 
That result bounds how often a rule must abstain; it does not tell a deployment what to do with an individual question, because the sampling process that produced the batch is outside its scope. 
Our refusal output is aimed at exactly that gap: the refusal is justified by online evidence about the difficulty of the instance and by a finite budget, and it can therefore be acted on per question. 

Closest in intuition is agreement-based selective prediction \citep{selectiveagreement2026}, which reads cross-head agreement through a Lipschitz version space to obtain a certified selective prediction. 
It shares our consensus-to-certificate move, but it certifies a fixed, one-shot decision, whereas we certify a stopping time under a budget and use time-uniform confidence sequences in place of version spaces. 
A second boundary is worth stating explicitly, because it is easy to overstate: certified self-consistency \citep{certifiedsc2025} also attaches concentration guarantees to majority voting, with bounds that hold uniformly over the sample size. 
The difference lies in what is certified and in how the budget enters: our rule is asked, for a fixed budget and target error, either to answer, to stop early, or to refuse with a bounded mis-refusal probability, and the certified analysis covers all three outputs jointly, with the budget entering as a feasibility criterion rather than as a stopping time that is free to grow. 
The boundary between the two formulations is the point at which we would expect the two lines of work to be compared directly, and we flag it as such rather than claiming a separation that our experiments cannot demonstrate. 

\subsection{Bandits with abstention and best-arm identification}\label{sec:rel-bai}

Our work is theoretically adjacent to bandits with abstention and to best-arm identification (BAI). 
Bandits with abstention \citep{banditabstention2024,banditabstention2,riskahabstain2025} and Bayesian BAI with abstention \citep{bayesianba2026} optimize regret or budget allocation over an action space the learner chooses from, rather than certifying the output of a single decision layer over a fixed pool. 
Classical BAI \citep{kaufmann2016,garivier2016,kalyanakrishnan2012,jamieson2016,bubeck2012} identifies the best arm in fixed-confidence or fixed-budget regimes with sample-complexity guarantees but no refusal output, since an arm must always be recommended. 
The distinction from our formulation is the output alphabet: BARS differs in treating refusal as a third output with a bounded mis-refusal probability, a formulation that fixed-confidence best-arm identification does not consider because it has no budget under which answering could be infeasible. 
The closest formulation on the impossibility side is the conformal impossibility result discussed above \citep{crc2026}, which bounds abstention from outside the sampling process; we bound it from inside, conditionally on the observed evidence. 

\subsection{Positioning along five dimensions}\label{sec:rel-table}

Table~\ref{tab:neighbours} places the four nearest families and BARS on the same five axes. 
The axes are chosen so that each one corresponds to a place where a deployment decision is made: what the answer space is, what the rule is allowed to output, whether the budget is a hard constraint, what the guarantee is about, and whether the refusal rate is bounded. 

\begin{table*}[t]
\caption{Nearest families along five dimensions. ``Hard budget'' means that the total number of samples is bounded by a constant fixed before the run; ``refusal bounded'' means that the probability of refusing an instance the budget could have handled carries a stated bound.}
\label{tab:neighbours}
\small
\centering
\begin{tabular}{@{}>{\raggedright\arraybackslash}p{0.20\textwidth}>{\raggedright\arraybackslash}p{0.14\textwidth}>{\raggedright\arraybackslash}p{0.16\textwidth}>{\raggedright\arraybackslash}p{0.14\textwidth}>{\raggedright\arraybackslash}p{0.20\textwidth}@{}}
\toprule
Family & Answer space & Outputs & Hard budget & Guarantee, and on what \\
\midrule
Self-consistency and its adaptive variants \citep{wang2023selfconsistency,esc2024,cges2025,dase2026,mars2026} & discrete & answer & no (stop rule free to grow) & empirical savings; heuristic margin criteria \\
Conformal abstention \citep{conformalabstention2024,sconu2025,crc2026} & fixed batch & answer, abstain & not applicable & distribution-free risk and abstention bounds for a one-shot decision \\
Agreement-based selective prediction \citep{selectiveagreement2026} & fixed features & answer, abstain & not applicable & certified selective risk in a version space \\
Certified self-consistency \citep{certifiedsc2025} & discrete & answer, continue & no & concentration bound uniform over sample size \\
Best-arm identification \citep{kaufmann2016,garivier2016} & discrete arms & arm & fixed-confidence or fixed-budget & sample complexity; no refusal \\
\midrule
BARS (this paper) & unknown discrete & early stop, answer, refuse & yes ($E^*$) & joint answering-error and mis-refusal bounds (certified mode), plus a two-point lower bound \\
\bottomrule
\end{tabular}
\end{table*}

\section{Problem formulation and the BARS algorithm}\label{sec:method}

\subsection{The decision problem, its outputs and its error quantities}\label{sec:method-formulation}

We formulate the deployment problem as identifying the most probable color of an unknown distribution under a sampling budget and an error-rate target. 

\begin{center}
\begin{minipage}{0.92\columnwidth}
\textbf{Problem (budgeted majority identification with refusal).} 
We are given an unknown distribution $p$ over an unknown set of colors, and we may draw iid samples from $p$, each sample consuming one unit of a total budget $E^*$. 
The rule must return one of three outputs: an answer (a color), an early stop with an answer that uses fewer than $E^*$ samples, or an explicit refusal. 
Conditional on answering, the probability of returning a color other than the true majority must be at most $p^*$. 
When answering is in fact possible within the budget, the probability of refusing must be at most $\delta$. 
\end{minipage}
\end{center}

The budget is a per-path cap rather than an expected-value constraint: no single execution of the rule draws more than $E^*$ samples, and we report the expected sample count because the quantity of interest in deployment is the average cost over a batch of questions. 
The only assumptions are that a unique majority exists, that is $\Delta = p_{(1)} - p_{(2)} > 0$, and that the samples are iid. 
The number of colors $K$ and the shape of $p$ are otherwise unrestricted, and we write $\hat K_t$ for the number of distinct colors observed up to time $t$. 

\begin{assumption}[Bounded number of colors]\label{asm:color-bound}
A known upper bound $\bar K \ge K$ on the number of colors is available to the rule; the certified width (Section~\ref{sec:method-width}) and the feasibility criterion of the naive branch (Section~\ref{sec:method-branches}) use $\bar K$ in place of the unknown $K$. 
\end{assumption}

Two events recur throughout the analysis, and we fix their notation now. 
Let $N = E^* - \tau$ denote the remaining budget at the branch-entry time $\tau$. 
The \emph{naive-feasibility event} $\mathcal{A}$ holds when the Hoeffding majority rule of the naive branch meets the target with the remaining budget:
\[
\mathcal{A} \;:=\; \left\{ (K-1)\, e^{-N \Delta^2 / 2} \le p^* \right\},
\]
evaluated with the true gap $\Delta$ and the remaining budget $N$ at the branch entry; it is an analyst-level notion that the algorithm never observes. 
Because $\mathcal{A}$ is defined through the Hoeffding rule, the mis-refusal guarantee is relative to that rule and not to the best possible rule. 
The \emph{certification event} $\mathcal{I}$ holds when the conservative evidence --- the plug-in lower endpoint $d_{\mathrm{use}}$ and, in the certified mode, the color-count bound $\bar K$ --- deems the target infeasible:
\[
\mathcal{I} \;:=\; \left\{ (\bar K - 1)\, e^{-N d_{\mathrm{use}}^2 / 2} > p^* \right\}.
\]
A \emph{mis-refusal} is the joint event $\mathcal{A} \wedge \text{refusal}$: refusing although the naive Hoeffding rule would have sufficed. 
The rule can refuse only through the naive branch, so that a refusal is the event $\{\text{enter the naive branch}\} \wedge \mathcal{I}$; instances that enter the fast branch are answered. 
The label ``infeasible within budget'' printed by the rule is therefore a rule-specific statement, not a claim that no rule could answer. 
Keeping them apart is the reason the paper distinguishes the certified and practical modes throughout: the two modes do not differ in the rule, only in which of these three quantities is bounded. 


\subsection{Notation}\label{sec:method-notation}

All symbols are collected in Table~\ref{tab:symbols}. 
We distinguish the true number of colors $K$, its online estimate $\hat K_t$ and the bound $\bar K$ required by the certified mode, and we use $w$ for the run length of the fast branch and $w_t$ for the confidence half-width of the exploration phase; the two are never interchanged. 

\begin{table*}[t]
\caption{Notation. $p$ denotes the true distribution, with probabilities arranged in descending order; $\hat{\cdot}$ denotes the online estimate of the corresponding quantity, and $\bar{\cdot}$ a known bound used by the certified mode.}
\label{tab:symbols}
\small
\centering
\begin{tabular}{@{}lll@{}}
\toprule
Symbol & Meaning & Notes \\
\midrule
$p,\ p_{(1)}>p_{(2)}\ge\cdots$ & true distribution (desc.) & $p_{(1)}$ is the majority \\
$K$ & number of colors (unknown) & estimated online by $\hat K_t$ \\
$\bar K$ & known upper bound on $K$ & required by the certified mode only (Assumption~\ref{asm:color-bound}) \\
$\Delta = p_{(1)}-p_{(2)}$ & top-two gap, the difficulty of the instance & larger is easier \\
$\sigma^2$ & per-sample variance of the gap & $p_{(1)}(1-p_{(1)})+p_{(2)}(1-p_{(2)})+2p_{(1)}p_{(2)}$ \\
$E^*$ & sampling budget & at most this many samples \\
$p^*$ & target conditional error rate & bound on the error of an answered question \\
$\delta$ & mis-trigger bound & nominal level of the certified width \\
$B_{\mathrm{plug}}$ & plug-in bias of the refusal decision & no distribution-free bound; calibrated empirically \\
$\Delta_{\mathrm{high}}$ & fast-branch trigger threshold & lower-bound claim on $\Delta$ \\
$T_{\mathrm{cap}}$ & exploration cap & bounds the exploration phase \\
$\mathrm{min\_t}$ & minimum sample count before triggering & safeguard against degenerate triggers \\
$\hat p_t,\ \hat\Delta_t$ & empirical frequencies and gap at $t$ & maintained online \\
$\hat\sigma_t^2$ & plug-in variance of the gap & see the formula above \\
$w_t$ & confidence half-width at $t$ & certified or practical mode \\
$\tau$ & samples used at branch entry & trigger time \\
$w$ & run length of the fast branch & chosen by the dynamic program \\
$N = E^*-\tau$ & remaining budget & shared by the fast and naive branches \\
$d_{\mathrm{use}}$ & conservative lower endpoint of $\Delta$ & $\max(0,\hat\Delta_\tau-w_\tau)$, used in the feasibility criterion \\
$p_{\mathrm{err}}[w;q]$ & error of the branch rule under candidate $q$ & exact chain value; see Section~\ref{sec:method-branches} \\
$\rho_N$ & folding residual of the pessimistic chain & $(\bar K-1)e^{-Nd_{\mathrm{use}}^2/8}$ \\
$\mathcal{A},\ \mathcal{I}$ & naive-feasibility and certification events & analyst level; defined in Section~\ref{sec:method-formulation} \\
\bottomrule
\end{tabular}
\end{table*}

\subsection{The rule}\label{sec:method-algorithm}

BARS alternates between exploration and one of three decision branches, and the pseudocode is given in Algorithm~\ref{alg:main}. 
Sections~\ref{sec:method-width} and~\ref{sec:method-branches} state the two ingredients it relies on. 
We describe the rule as a sequence of decisions rather than as three phases, because the three outputs are mutually exclusive at the level of a single question: once the trigger fires, the naive and refusal branches are unreachable for that question. 

\begin{algorithm*}[t]
\caption{BARS: budget-aware adaptive majority voting with refusal}
\label{alg:main}
\small
\begin{algorithmic}[1]
\Require{budget $E^*$, error $p^*$, threshold $\Delta_{\mathrm{high}}$, confidence $\delta$, cap $T_{\mathrm{cap}}$, $\mathrm{min\_t}$, width mode (certified or practical)}
\Require{Choose-$w$ mode (pessimistic or optimistic)}
\State \textbf{Initialize} counts $\leftarrow \{\}$, $t \leftarrow 0$, run $\leftarrow 0$, last $\leftarrow \emptyset$ \Comment{$t$: samples used}
\Repeat
  \State $c \leftarrow \textsc{Sampling}()$ \Comment{drawing a ball from the bag (one LLM decode)}
  \State $t \leftarrow t+1$; counts$[c] \leftarrow$ counts$[c] + 1$ \Comment{update counts}
  \State $(\hat p_t, \hat\Delta_t, \hat\sigma_t^2) \leftarrow \textsc{Top2Stats}(\text{counts}, t)$ \Comment{online estimates}
  \State $w_t \leftarrow \textsc{Width}(\hat\sigma_t^2, t)$ \Comment{certified or practical}
  \If{$t \ge \mathrm{min\_t}$ \textbf{and} $\hat\Delta_t - w_t \ge \Delta_{\mathrm{high}}$}
    \State $\tau \leftarrow t$; \textbf{goto} \textsc{FastBranch} \Comment{enter the fast branch}
  \EndIf
  \If{$t = T_{\mathrm{cap}}$}
    \State $\tau \leftarrow t$; \textbf{goto} \textsc{NaiveBranch}
  \EndIf
\Until{$\text{false}$}
\Statex
\Function{FastBranch}{}
  \State $N \leftarrow E^*-\tau$ \Comment{remaining budget}
  \State $w \leftarrow \textsc{ChooseW}(\hat p_\tau, N, p^*, \text{mode})$ \Comment{dynamic program; mode is pessimistic or optimistic}
  \If{$w \neq \textsc{None}$}
    \Repeat
      \State $c \leftarrow \textsc{Sampling}()$; $t \leftarrow t+1$; counts$[c] \leftarrow$ counts$[c]+1$
      \State run $\leftarrow (c = \text{last}) ?\ \text{run}+1 : 1$; last $\leftarrow c$
      \If{run $\ge w$} \State \textbf{break} \EndIf
    \Until{$t \ge E^*$}
    \State \Return $\textsc{Majority}(\text{counts})$ \Comment{majority over the full trajectory}
  \Else
    \State \textbf{goto} \textsc{NaiveBranch} \Comment{no admissible window within budget: fall back to the feasibility criterion}
  \EndIf
\EndFunction
\Statex
\Function{NaiveBranch}{}
  \State $N \leftarrow E^* - \tau$ \Comment{remaining budget; set here as well, because the cap can route directly to this branch}
  \State $d_{\mathrm{use}} \leftarrow \max(0,\ \hat\Delta_\tau - w_\tau)$ \Comment{conservative lower endpoint of $\Delta$}
  \State $K_{\mathrm{crit}} \leftarrow \hat K_\tau$ \Comment{observed distinct colors}
  \If{\textbf{certified mode}} \State $K_{\mathrm{crit}} \leftarrow \bar K$ \Comment{use the Assumption~\ref{asm:color-bound} bound, as the theorems do} \EndIf
  \If{$(K_{\mathrm{crit}} - 1)\,e^{-N\, d_{\mathrm{use}}^2/2} \le p^*$}
    \State take $N$ more balls; \Return $\textsc{Majority}(\text{counts})$
  \Else
    \State \Return $\textsc{Refuse}$ \Comment{infeasible within budget}
  \EndIf
\EndFunction
\end{algorithmic}
\end{algorithm*}

\subsection{Exploration and the two width modes}\label{sec:method-width}

The validity of the trigger $\hat\Delta_t - w_t \ge \Delta_{\mathrm{high}}$ depends on whether $w_t$ is a valid time-uniform lower confidence half-width for the true gap $\Delta$. 
We provide two choices, and the choice is the only place in the rule where a guarantee is traded for a higher answer rate. 

\begin{itemize}
  \item \textbf{Certified mode.} 
  With $\bar K$ as in Assumption~\ref{asm:color-bound} --- in the LLM setting one may take the size of the parsed answer space --- set
  \[
  w_t = 2\sqrt{\frac{2\bar\sigma_t^2\,(2\ln t + \ln\frac{\bar K\pi^2}{3\delta})}{t}},
  \]
  where $\bar\sigma_t^2$ is the plug-in variance, floored as described below. 
  The leading factor $2$ absorbs the Lipschitz constant of the top-two gap functional, and the $\ln \bar K$ term pays for a union bound over colors; both are made explicit so that $w_t$ is a time-uniform lower confidence half-width for the true $\Delta$: on the good event $\mathcal{G}$ of probability at least $1-\delta$, $|\hat\Delta_t - \Delta| \le w_t$ holds simultaneously over the exploration times (Corollary~\ref{cor:trigger}). 
  Because the plug-in variance of a batch of identical draws is exactly zero, the variance used in $w_t$ is floored at a positive value $\sigma_{\min}^2 > 0$ that decays with $t$; without the floor the width would vanish on the first samples and the trigger would fire on no evidence (Section~\ref{sec:theory-tools}, Remark~\ref{rem:floor}). 
  The drawbacks are that its constants are conservative and that the \emph{certification time}
\[
T_{\mathrm{cert}} \;:=\; \min\big\{t \ge \mathrm{min\_t}:\ w_t \le \hat\Delta_t - \Delta_{\mathrm{high}}\big\},
\]
the smallest $t$ at which the certified trigger can fire, scales roughly as $(\Delta-\Delta_{\mathrm{high}})^{-2}$; small budgets are therefore pushed towards refusal. 
  Section~\ref{sec:experiments-cert} reports its explicit values. 
  \item \textbf{Practical mode.} 
  \[
  w_t = z\sqrt{\hat\sigma_t^2 / t}, \quad \text{typically } z=2.
  \]
  This is a heuristic, variance-scaled width: it controls the deviation only at the current sample size and carries no time-uniform guarantee, so its safety is calibrated empirically rather than proved. 
  Experiments indicate that $z \ge 3$ substantially mitigates mis-triggers on hard instances; we recommend $z = 2$ for interactive deployment, and $z \ge 3$ when stricter mis-trigger control is required (quantified in Section~\ref{sec:experiments-tuning}). 
\end{itemize}

The certified mode is used when an auditable certificate is required; the practical mode is the configuration we expect an operator of a small sampling budget to run. 

\subsection{The algorithm stages}\label{sec:method-branches}

\paragraph{Exploration.} 
As samples arrive, the rule updates the empirical frequencies $\hat p_t$, the top-two gap $\hat\Delta_t$ and the plug-in variance $\hat\sigma_t^2$, tests the trigger $\hat\Delta_t - w_t \ge \Delta_{\mathrm{high}}$ and, if the trigger has not fired by $t = T_{\mathrm{cap}}$, leaves the exploration phase. 
The trigger has two safeguards: $\mathrm{min\_t}$ prevents it from firing before any evidence on the variance exists, and $\Delta_{\mathrm{high}}$ requires the certified lower endpoint to clear a strictly positive threshold. 

\paragraph{The fast branch.} 
For a fixed run length $w$ and a candidate distribution $q$ over the observed colors, the dynamic program evaluates the exact error probability of the branch's answering rule,
\[
\begin{aligned}
p_{\mathrm{err}}[w; q] &= \Prob_{q,\,N}(\text{trajectory-majority error at the stop})\\
&= \big(U(w; q) - W(w; q)\big) + V(w; q) + F(w; q),
\end{aligned}
\]
where $U$ is the probability that the first $w$-run is completed by a non-majority color (\emph{run part}); $W$ the probability of that event with the majority still leading at the stop, in which case the trajectory-majority answer is correct; $V$ the probability that the majority completes the first $w$-run but does not lead at the stop (\emph{count-gap part}, which earlier analyses of run-based stopping omitted); $X$ the probability that the majority completes the run while still leading, so that stopping on a run has total probability $U+V+X$; and $F$ the probability that no color completes a run within $N$ samples and the fallback majority is wrong. 
Since $W \ge 0$, summing only $U+V+F$ gives a conservative upper bound, whereas the chain computes the exact value $U-W+V+F$; the construction of the chain, its state count and the residual that the pessimistic mode incurs are given in Appendix~\ref{app:dp}. 
The chosen run length is the smallest admissible $w$ for which the worst-case error rate, residual included, is at most $p^*$, and the branch stops at the first time the last $w$ samples share a color and answers with the majority of the full trajectory. 
The \emph{pessimistic} mode evaluates $p_{\mathrm{err}}$ at the distribution $q^{\mathrm{pess}}$ that pushes the majority probability down to its confidence-lower endpoint $\hat p_{(1)} - w_\tau$ and merges the residual mass into a single filler color; the \emph{optimistic} mode evaluates it at the plug-in distribution $\hat p_\tau$. 
The theorems of Section~\ref{sec:theory} hold for the pessimistic mode; the optimistic mode is an empirically validated relaxation. 

\paragraph{The naive branch and refusal.} 
With $d_{\mathrm{use}} = \max(0, \hat\Delta_\tau - w_\tau)$ as the conservative lower endpoint of $\Delta$ (the certified width already absorbs the factor-$2$ and $\ln \bar K$ constants), the branch checks $(K_{\mathrm{crit}}-1)e^{-N d_{\mathrm{use}}^2/2} \le p^*$, where $K_{\mathrm{crit}} = \hat K_\tau$ in the practical mode and $K_{\mathrm{crit}} = \bar K$ in the certified mode. 
If the criterion holds, the branch spends the remaining budget and takes the majority, and its error is controlled by the Hoeffding bound of Section~\ref{sec:theory-naive}; otherwise it refuses. 
The naive branch is retained even though our experiments never reach it, because the trigger and the criterion are logically independent: a question that fails the trigger can still be feasible, and a rule that dropped this branch would have no path for such questions. 

\subsection{Cost model}\label{sec:method-complexity}

All decision logic runs on the CPU: \textsc{Top2Stats} sorts in $O(\hat K_t \log \hat K_t)$, the width costs $O(1)$, and the dynamic program has $O(\hat K_\tau w)$ states for the run part plus $O(wN)$ for the count-gap tracking (Appendix~\ref{app:dp}). 
For language models the dominant cost is the number of forward passes, so the sample count is the cost metric throughout this paper.

\section{Theory}\label{sec:theory}

The results are ordered as they depend on one another, so that each one uses only what precedes it. 
The chain is: simultaneous coverage of the estimated gap, which certifies the trigger; the transfer of the fast-branch analysis to the worst-case distribution inside the confidence set, which certifies the fast branch; the two bounds that constitute the main deliverable, on answering and on refusing; the cost decomposition, which is what makes the rule budget-aware; and a lower bound that says how much of that cost is unavoidable. 
Proofs are collected in Appendix~\ref{app:proofs}. 

\subsection{Simultaneous coverage and the certified trigger}\label{sec:theory-tools}

We build on two classical tools. 
Hoeffding's inequality controls the deviation of an empirical mean from its expectation \citep{hoeffding1963}, and an empirical-Bernstein time-uniform confidence sequence controls the deviation \emph{simultaneously over all times} \citep{maurer2009,howard2021,waudbysmith2023}; the second is the direct source of the certified width. 

\begin{lemma}[Time-uniform confidence sequence, with the additive term retained]\label{lem:cs}
Let $X_t$ be iid and bounded in $[0,1]$, with mean $\mu$ and variance $\sigma^2$, let $\hat\sigma_t^2$ be the empirical variance, and let $c_t := 2\ln t + \ln\frac{\pi^2}{3\delta}$. 
For every exploration time $t \ge 2$, define
\[
w_t \;=\; \sqrt{\frac{2\hat\sigma_t^2\,c_t}{t}} \;+\; \frac{7\,c_t}{3\,t}.
\]
Then
\[
\Prob\Big(\exists\, t \ge 2:\ |\bar X_t - \mu| \ge w_t\Big) \;\le\; \delta .
\]
\end{lemma}

\begin{remark}[Why the additive term is not optional]\label{rem:floor}
The width has two terms and both are needed. 
The square-root term alone vanishes when the first draws coincide, because the plug-in variance of a constant batch is exactly zero; a rule that used it alone would certify a gap from no evidence at all. 
The additive term $7c_t/(3t)$ never vanishes, and it is the term that keeps the certified trigger valid on the first samples. 
Two conventions are in use in the literature and they differ here by a factor we quantify, because the difference matters for the reported budgets: the alternative convention absorbs the additive term into the multiplicative constant, which is safe only while the plug-in variance is bounded away from zero. 
The implementation and the figures of Section~\ref{sec:experiments-cert} use that second convention. 
At the exploration times reached there, the additive term is $5.7\%$, $17.7\%$ and $67.6\%$ of the square-root term for $\Delta = 0.2$, $0.4$ and $0.8$, so retaining it multiplies the certification time by about $1.1$, $1.4$ and $3.2$ respectively, while leaving the $(\Delta-\Deltahigh)^{-2}$ scaling unchanged. 
We report both sets of values in Section~\ref{sec:experiments-cert}.
The simulations confirm that the degenerate regime is not a corner case: with $\mathrm{min\_t}=2$ the square-root-only width certifies a gap on instances with $\Delta<\Deltahigh$ in $32.6$--$32.9\%$ of runs, against $0$ of $200{,}000$ for the two-term width (Section~\ref{sec:experiments-theorem-config}). 
\end{remark}

The details of the variance endpoint, the constants and the union bound over colors are in Appendix~\ref{app:proof-cs}. 
Because the order statistics $p_{(1)} \ge p_{(2)} \ge \cdots$ are $1$-Lipschitz in the $\ell_\infty$ sense, the same concentration bound transfers to the empirical top-two gap: $|\hat\Delta_t - \Delta| \le 2\max_c |\hat p_c - p_c|$. 
Combining the two with a union bound over the at most $\bar K$ colors yields the certified trigger used by the rule. 

\begin{corollary}[Certified trigger]\label{cor:trigger}
Let $\bar K$ be the known upper bound on the number of colors (Assumption~\ref{asm:color-bound}), and let
\[
w_t = 2\left[\sqrt{\frac{2\hat\sigma_t^2\,\bar c_t}{t}} + \frac{7\,\bar c_t}{3\,t}\right],
\qquad \bar c_t := 2\ln t + \ln\frac{\bar K\pi^2}{3\delta}.
\]
Then, on the event $\Gcal$ of probability at least $1-\delta$,
\[
\Prob\Big(\exists\, t \ge 2:\ |\hat\Delta_t - \Delta| \ge w_t\Big) \;\le\; \delta,
\]
and consequently, on $\Gcal$, the trigger $\hat\Delta_t - w_t \ge \Deltahigh$ certifies $\Delta \ge \Deltahigh$; the factor $2$ pays for the Lipschitz constant and $\ln\bar K$ for the union bound over colors. 
\end{corollary}

\subsection{The fast branch}\label{sec:theory-fast}

Entering the fast branch does not give the rule the true distribution, only a confidence set of radius $w_\tau$ around the empirical distribution. 
Our analysis therefore evaluates the branch's error at the most adversarial distribution inside that set, obtained by a pessimistic redistribution in which the majority probability is pushed to its minimum and the residual mass is merged into a single filler color. 
The next lemma states the dominance relation that makes this reduction legitimate, and it is why the dynamic program can be evaluated at a single distribution. 

\begin{lemma}[Pessimistic dominance]\label{lem:dom}
{\setlength{\emergencystretch}{1.5em}%
Fix the confidence configuration $(\hat p_\tau, w_\tau)$ at branch-entry time $\tau$, let $m = \hat p_{(1)} - w_\tau$ be the conservative lower endpoint of the majority probability, let $q^{\mathrm{pess}}$ be the folded two-color distribution whose majority probability is $m$ and whose remaining mass sits on a single filler color, and let $N = E^* - \tau$.\par} 
Then for every distribution $p$ with $p_{(1)} \ge m$ --- in particular every $p$ in the confidence set on $\Gcal$ --- every run length $w \ge 1$ and the same initial counts,
\[
\begin{aligned}
&\Prob_{p,\,N}\big(\text{trajectory-majority error at the branch stop}\big)\\
&\qquad\le\; p_{\mathrm{err}}[w; q^{\mathrm{pess}}] + \rho_N,
\end{aligned}
\]
where $p_{\mathrm{err}}$ is the exact chain value of Section~\ref{sec:method-branches} and $\rho_N = (\bar K - 1)e^{-N d_{\mathrm{use}}^2/8}$ is the folding residual of Appendix~\ref{app:proof-dom}. 
\end{lemma}

The residual is the price of the folding and it is small for the same reason that the problem is easy: on the paths where the folded process is killed by a filler run while the majority still leads, the unfolded process continues, and it can err only if some minority color overtakes the majority against a negative drift. 
This is a large-deviation event: on $\Gcal$ the Hoeffding inequality combined with the reflection principle bounds its probability by $(\bar K-1)e^{-N\Delta^2/8} \le \rho_N$, where $d_{\mathrm{use}} \le \Delta$ is used. 
On the tested confidence-set grid this residual is numerically below $10^{-5}$ (Section~\ref{sec:experiments-benefit}). 

\begin{theorem}[Fast-branch error, certified mode and pessimistic chooser]\label{thm:fast}
Run the rule with the certified width of Corollary~\ref{cor:trigger} and the pessimistic Choose-$w$. 
Conditioned on the good event $\Gcal$, if the dynamic program returns a run length $w$ with $p_{\mathrm{err}}[w; q^{\mathrm{pess}}] + \rho_N \le p^*$, then
\[
\Prob\big(\text{wrong answer} \mid \text{enter the fast branch},\ \Gcal\big) \;\le\; p^* ,
\]
because the error event at the stop is the union of three disjoint events that the chain enumerates: a minority completes the first $w$-run and leads at the stop, the majority completes it without leading, or no run completes within $N$ samples and the branch falls back to the trajectory majority, the last being controlled by the majority Hoeffding bound of Theorem~\ref{thm:error}. 
The criterion $p_{\mathrm{err}} + \rho_N \le p^*$ charges the residual to the same $p^*$ budget; numerically $\rho_N \le 10^{-5}$, so the chooser's output is unaffected. 
\end{theorem}

The fast branch answers with the majority of the whole trajectory rather than with the color that completed the run, and the difference can be quantified exactly. 

\begin{proposition}[Trajectory-majority benefit]\label{prp:benefit}
Consider the fast branch entered with run length $w$ under $\Gcal$, and let $U, W, V, X$ be the chain quantities of Section~\ref{sec:method-branches} evaluated at $q^{\mathrm{pess}}$. 
Conditional on a run completing, the run-color rule errs on the mass $U$ and the trajectory-majority rule on the mass $(U - W) + V$, so the two error probabilities differ by exactly
\[
P_{\mathrm{err}}^{\mathrm{run}} - P_{\mathrm{err}}^{\mathrm{traj}} \;=\; \frac{W - V}{U + V + X},
\]
where $U + V + X = \Prob(\text{the branch stops at a } w\text{-run})$: $U$ has a minority completing the run and leading at the stop, $V$ has the majority completing the run without leading, and $X$ has the majority completing it while still leading. 
The sign of $W - V$ is not fixed by the identity: the trajectory majority is better exactly when $W > V$, and Section~\ref{sec:experiments-benefit} reports that $W - V$ was non-negative at every point of the tested grid, with the gap growing in the remaining budget $N$. 
\end{proposition}

\subsection{The naive branch, refusal, and the main guarantees}\label{sec:theory-naive}

After $N$ iid samples the majority is not the true majority with probability at most $(\bar K-1)e^{-N\Delta^2/2}$, by Hoeffding's inequality and a union bound over the at most $\bar K$ colors. 
This bound is what the feasibility criterion of the naive branch evaluates at the conservative endpoint $d_{\mathrm{use}}$ instead of at $\Delta$. 

\begin{theorem}[Joint answering and mis-refusal guarantees, certified mode and pessimistic chooser]\label{thm:error}
Run the rule with the certified width of Corollary~\ref{cor:trigger} and the pessimistic Choose-$w$. 
Then, for any true distribution $p$ with $\Delta > 0$,
\[
\begin{aligned}
\underbrace{\Prob(\text{wrong majority} \wedge \text{answers})}_{\text{answering error (joint)}} &\le p^* + \delta,\\
\underbrace{\Prob(\text{refuses} \wedge \mathcal{A})}_{\text{mis-refusal (joint)}} &\le \delta + B_{\mathrm{plug}},
\end{aligned}
\]
where $\mathcal{A}$ is the naive-feasibility event of Section~\ref{sec:method-formulation} and
\[
B_{\mathrm{plug}} \;=\; \Prob\big(\text{refuses} \wedge \mathcal{A} \wedge \Gcal\big)
\]
is the plug-in bias: it is nonzero only when the conservative endpoint $d_{\mathrm{use}}$ falls short of the true $\Delta$ far enough that the criterion refuses an instance the budget would have handled. 
\end{theorem}

Three remarks delimit what Theorem~\ref{thm:error} guarantees and what does not. 
First, the joint form is the natural statement for guarantees that involve the observed events ``answers'' and ``refuses''; conditioning them out gives $(p^* + \delta)/\Prob(\text{answers})$ and $(\delta + B_{\mathrm{plug}})/\Prob(\mathcal{A})$ respectively, which reduce to the joint bounds when the answer rate, or the feasibility rate, is close to one. 
We therefore quote the joint form here and the conditional numbers in the experiments, and we never convert between them implicitly. 
Second, the sufficient condition for $B_{\mathrm{plug}} = 0$ is $d_{\mathrm{use}} = \Delta$ together with $K = \bar K$; when $K < \bar K$ the conservatism introduced by the color-count bound is a further source of mis-refusal that the plug-in bias alone does not capture. 
Third, $B_{\mathrm{plug}}$ has no distribution-free upper bound, and we do not claim one: it is calibrated empirically in Section~\ref{sec:experiments}, and the same statement is why the certified mode is not offered as a guarantee of the practical one. 
Finally, the guarantee of the practical width mode is empirical rather than a consequence of Theorem~\ref{thm:error}; the two modes differ in the width alone.

\subsection{Expected sampling cost}\label{sec:theory-expected}

For a Bernoulli sequence with success probability $q$, the expected number of trials until the first run of $w$ consecutive successes is $(1-q^w)/\big((1-q)q^w\big)$ \citep{feller1968}. 
This identity is what makes the cost of the fast branch predictable in closed form. 

\begin{theorem}[Expected-samples decomposition]\label{thm:expected}
Let $P_{\mathrm F}$, $P_{\mathrm N}$ and $P_{\mathrm R}$ denote the fast, naive and refusal branch probabilities, and let $\tau_{\mathrm F}$ and $\tau_{\mathrm R}$ denote the fast-branch trigger time and the refusal decision time; here $\tau_{\mathrm R} = T_{\mathrm{cap}}$ when the branch is entered from the main loop, and $\tau_{\mathrm R} = \tau_{\mathrm F}$ when the fast branch falls back. 
Write $T_{\mathrm{run}}(\hat p_{(1)}, w)$ for the run waiting time with Bernoulli parameter equal to the plug-in majority probability at branch entry. 
Then
\[
\begin{aligned}
\E[\text{total samples}] &= P_{\mathrm F}\cdot\big(\E[\tau_{\mathrm F}] + \E[\min(T_{\mathrm{run}}(\hat p_{(1)}, w), N)]\big)\\
&\quad + P_{\mathrm N}\cdot E^* + P_{\mathrm R}\cdot \E[\tau_{\mathrm R}],
\end{aligned}
\]
with $N = E^* - \tau_{\mathrm F}$; the truncation accounts for the fact that the fast branch cannot run past the budget, and it is immaterial when $N$ exceeds the typical waiting time. 
The identity is exact when the two-color waiting time is used; with the true multi-color race the fast-branch term is an upper bound, and Monte Carlo verification matches the dynamic-programming output to $10^{-4}$. 
\end{theorem}

Because the decomposition is closed-form, the amortized cost of a batch can be computed before the batch is run, and a budget can be chosen to match a latency target. 
We do not use this bound to explain the measured sample counts, because it is conservative by construction: it charges the exploration cap and ignores the samples saved by early refusal. 

\subsection{Why small budgets force refusal: a lower bound}\label{sec:theory-lower}

The last result explains why refusal is not an implementation artefact but a consequence of the budget. 
We consider a two-point family of distributions $\{P^{+},P^{-}\}$ whose majorities are two different colors and whose top-two gap is the same $\Delta$. 
Their single-sample KL divergence is $\Delta\ln\frac{1+\Delta}{1-\Delta} = 2\Delta^2 + O(\Delta^4)$. 
By Le Cam's two-point method together with Stein's lemma \citep{tsybakov2009} we obtain the following bound. 

\begin{theorem}[Two-point lower bound]\label{thm:infolb}
Let $\delta \in (0,\tfrac14)$. 
Any sequential rule that outputs the correct majority with probability at least $1-\delta$ on each distribution of the two-point family satisfies, in expectation,
\[
n \;\ge\; c\,\Delta^{-2}\ln\frac{1}{\delta}
\]
for an absolute constant $c>0$. 
\end{theorem}
\section{Experiments}\label{sec:experiments}
The experiments divide into two settings and: random-seed synthetic instances whose generating distributions are known, where the predictions of the rule can be checked exactly, and a real model at a deployment-scale budget.
The two settings share one implementation of the rule and one set of competing rules.

\subsection{Experiments design}\label{sec:experiments-design}

\textbf{Competing rules.}
{\setlength{\emergencystretch}{1.5em}%
We compare against two families of competitors in Table~\ref{tab:neighbours}. 
The first family contains fixed-budget and fixed-confidence best-arm-identification (BAI) rules that are evaluated on the same sampled stream: the fixed-budget majority rule with the full budget (\textsc{Fixed-$N$}), Successive Rejects \citep{kaufmann2016}, Sequential Halving \citep{kalyanakrishnan2012}, LUCB \citep{kalyanakrishnan2012}, lil'UCB \citep{jamieson2016} and the Hoeffding race \citep{maron1993}. 
The second family contains the sequential early-stopping rules designed for LLM majority voting that we reviewed in Section~\ref{sec:rel-tts}: early-stopping self-consistency (ESC)~\citep{esc2024}, its difficulty-adaptive offspring~\citep{difficulty2024}, confidence-guided early stopping (CGES)~\citep{cges2025}, the sequential-evidence rule DASE~\citep{dase2026}, the margin-adversarial rule MARS~\citep{mars2026} and the martingale majority certificate of certified self-consistency~\citep{certifiedsc2025}. 
Each competitor is implemented from its published stopping rule, at the parameters its authors report as default, and every rule consumes the same sampled stream, uses the same answer normalisation and is scored against the same reference; the implementations and their exact stopping criteria are listed in Appendix~\ref{app:competitors}. 
Two of the six LLM competitors need information the pool-replay protocol does not carry: CGES scores answers by their decoding confidence and MARS needs partial-trace states. 
We therefore evaluate CGES in its confidence-free form, where the posterior over candidate answers is the Dirichlet-multinomial posterior implied by the counts, and MARS in its answer-level form, where the margin is required to exceed a conservative bound on future vote movement; both reductions are stated so that they are not mistaken for the full methods.\par} 

The \textsc{Fixed-$N$} rule and the BAI rules always return an answer, since a best arm is always recommended; BARS and the LLM early-stopping rules can refuse or fall back. 
A comparison must therefore separate two quantities that a single accuracy number would conflate: the error rate \emph{on the instances a rule answers}, and the rate at which a rule correctly declines instances for which no budget-feasible answer exists. 
We define the second quantity against an oracle that knows $p$: an instance is \emph{feasible} when the oracle's criterion $(K-1)e^{-N\Delta^2/2}\le p^*$ holds at the full budget, and we report the \emph{mis-refusal rate} (refusing a feasible instance) and the \emph{correct refusal rate} (refusing an infeasible one) separately. 
This is the synthetic counterpart of the selective-prediction convention of Section~\ref{sec:rel-conformal}, and it is the only protocol under which a rule that declines to answer is not penalised for a decision that the fixed-budget baselines are structurally unable to make. 

\textbf{Reporting rules.}
An \emph{answer rate} is the fraction of questions the rule answers, so that its complement is the refusal rate and the two sum to one, as in selective prediction. 
\emph{Accuracy on answered questions} divides by the number of answered questions, whereas the mean sample count divides by all questions of the pool and therefore includes the questions that were refused early; the $22.3\%$ figure below is of the second kind, and we state it with its denominator every time it is quoted. 
The synthetic cells report conditional error rates averaged over seeds, and the LLM cells report proportions with Wilson $95\%$ intervals, because the latter are binomial proportions over 80 questions and a standard deviation would misstate their uncertainty; the synthetic figures are exact up to the Monte Carlo verification of the dynamic program ($10^{-4}$) and the seed count. 
Competitor curves are drawn with the same seeds and the same instance batches as the BARS curves throughout, so every point in the synthetic figures is a paired comparison. 

\subsection{Random-seed synthetic experiments}\label{sec:experiments-synth}

Three properties make synthetic instances the right place to begin.
First, they cover the situations a deployed system actually meets: a grid of difficulties in the top-two gap $\Delta$ that runs from instances the budget settles easily to instances it cannot settle at all, crossed with budgets from tens to hundreds of samples, so that the rule is not tested only where it happens to be comfortable.
Second, the generating distribution is known, because it is the distribution we drew from: the oracle feasibility criterion $(K-1)e^{-N\Delta^2/2}\le p^*$, the true top-two gap $\Delta$ and the true majority are all exactly computable, and none of the three is available on a real dataset.
Third, and because of the second, the synthetic side is where the rule can be audited rather than merely scored: the certified mode's predictions can be checked against the quantity they predict, and the rule can be priced separately where it is right and where it is wrong.
Section~\ref{sec:experiments-llm} can do none of this; it answers a different question, namely whether the behavior survives contact with a real model.

\subsubsection{Instances and the synthetic parameter set}\label{sec:experiments-families-setup}

The instances are drawn from a $\Delta$-grid family: the two leading colors split $0.8$ of the probability mass with a gap of exactly $\Delta$, and the remaining eight colors share the residual $0.2$ equally, so that difficulty is the single axis along which the instances differ.
Default parameters are $E^*=200$, $p^*=0.05$, $\Delta_{\mathrm{high}}=0.1$, $\delta=0.05$, $T_{\mathrm{cap}}=150$ and $\mathrm{min\_t}=8$, with the practical width ($z=2$) and the optimistic chooser; the main cells use 2000 seeds and the tuning cells 600--1200.

\subsubsection{Behavioral signature across difficulty}\label{sec:experiments-sweep}

\begin{figure*}[htbp]
  \centering
  \includegraphics[width=\textwidth]{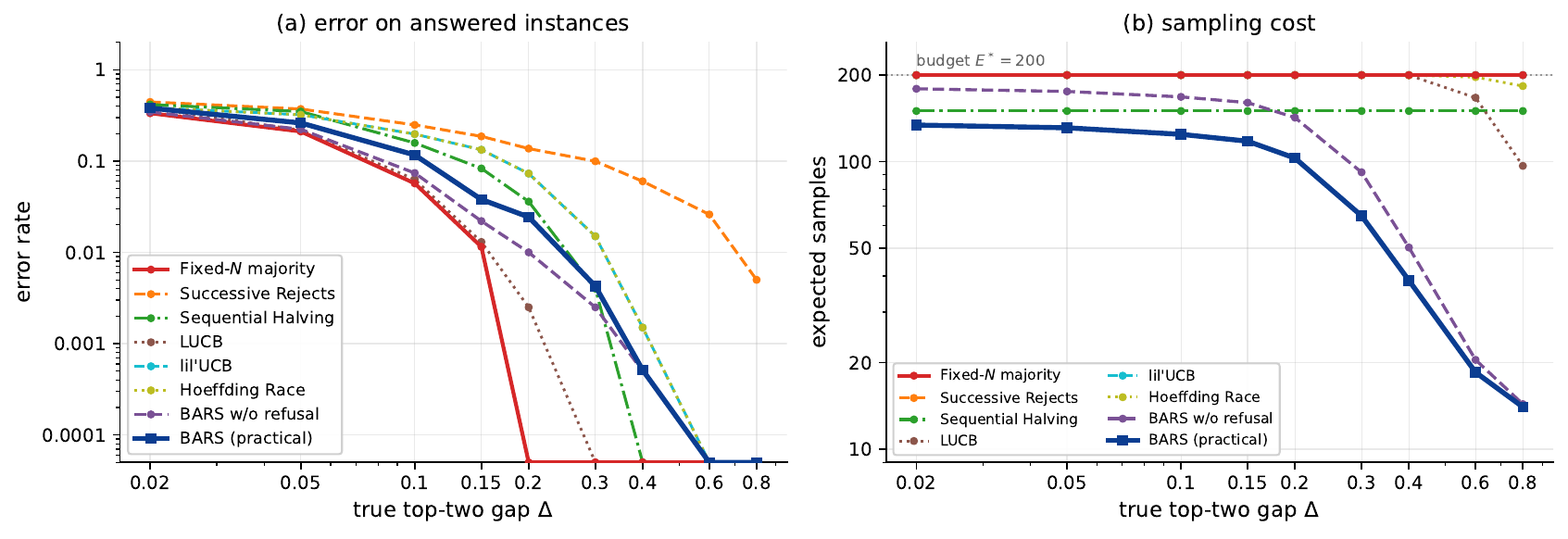}
  \caption{Results across the $\Delta$ sweep under the practical mode, with the competing rules on the same instances. Left: conditional error rate, that is the error over the instances each rule answers (the BAI rules and \textsc{Fixed-$N$} answer every instance, so their conditional and total error coincide). Right: expected sample count. Every curve is averaged over 2000 seeds per cell; the fixed-budget baseline draws exactly 200 samples on every instance.}
  \label{fig:delta-sweep}
\end{figure*}

Figure~\ref{fig:delta-sweep} shows how the practical mode behaves as the difficulty of the instance varies, and places it against the competitors on the same instances. 

At $\Delta = 0.4$ the fast branch handles $97.3\%$ of the seeds with a mean of $38.6$ samples, which is $19\%$ of the budget, and at $\Delta = 0.8$ every seed is answered with no error in $14$ samples on average, a $93\%$ reduction in sample count. 
The same two points cost the fixed-budget baseline its full $200$ samples, and cost LUCB $199.8$ and $96.7$ samples respectively, so the saving is a property of stopping rather than of the answer rule: all four rules agree on the answer at $\Delta = 0.8$, and the difference is entirely in how long they wait. 
For $\Delta \le 0.2$ the answer rate of BARS falls to $13$--$47\%$, so most hard instances are refused rather than guessed, and a fixed-majority baseline that draws the full $N=200$ samples still leaves $21$--$33\%$ error at $\Delta \le 0.05$. 
The elimination-style competitors do worse, not better, as the budget grows useless: Successive Rejects and Sequential Halving leave $35$--$44\%$ error at $\Delta \le 0.05$, because a fixed-budget BAI rule must spend its budget eliminating arms it cannot distinguish and then recommend one of them. 
Exhausting the budget does not rescue these instances, which is the regime the refusal branch exists for. 

\begin{figure*}[htbp]
  \centering
  \includegraphics[width=\textwidth]{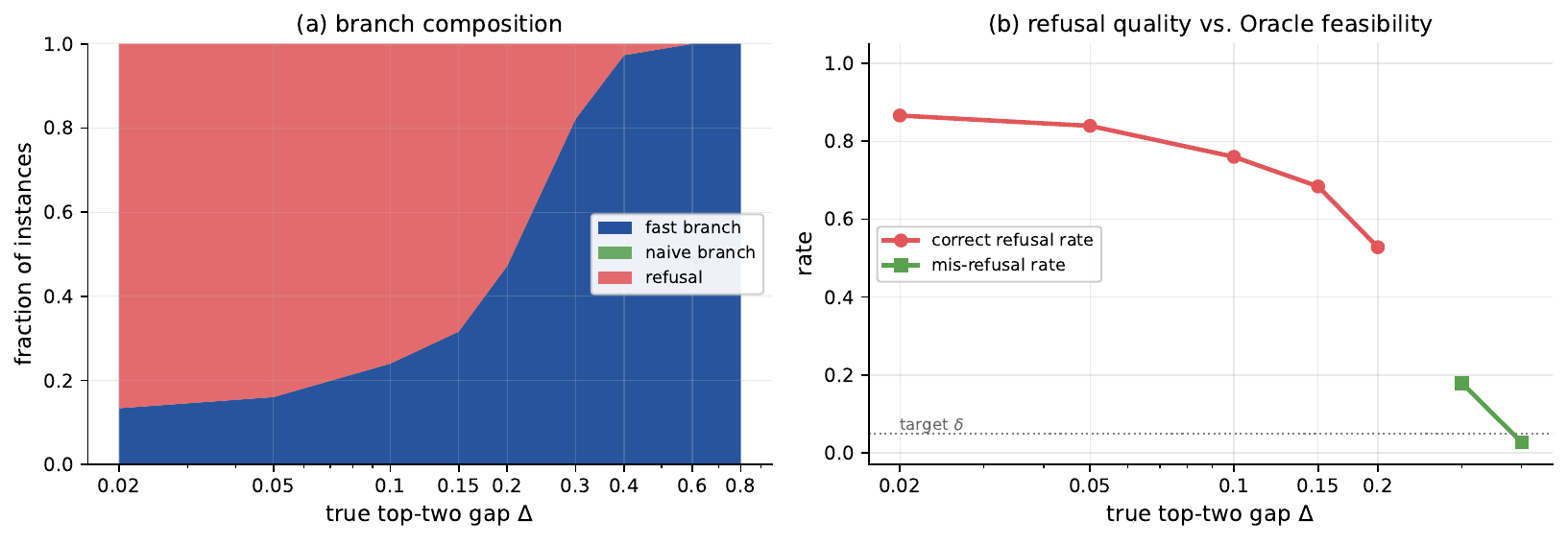}
  \caption{Branch composition and refusal quality as functions of $\Delta$. Left: the refusal region marks the instances for which the criterion cannot establish feasibility within the budget; the transition to the fast branch is consistent with the $(\Delta-\Delta_{\mathrm{high}})^{-2}$ certification cost of Section~\ref{sec:method-width}. Right: refusal correctness against the oracle feasibility criterion of Section~\ref{sec:experiments-design}, that is the fraction of infeasible instances that are refused and of feasible ones that are not.}
  \label{fig:branches}
\end{figure*}

The refusal decision itself is audited in Figure~\ref{fig:branches}. 

The composition in Figure~\ref{fig:branches} shows a single transition rather than a continuum: larger gaps are handled by the fast branch, smaller ones by refusal, and the boundary moves as the certification cost predicts. 
The detailed data sweep in numbers rule by rule, is shown in Table~\ref{tab:delta-sweep}.
On the right panel, the correct refusal rate is $0.53$ at $\Delta = 0.2$, $0.76$ at $\Delta = 0.1$ and $0.87$ at $\Delta = 0.02$, while the mis-refusal rate is below $3\%$ everywhere on the grid except at $\Delta = 0.3$, where it is $18\%$: the rule errs towards answering, not towards refusing, which is the direction the error budget $p^*$ is stated in. 
The na\"ive branch is never reached in this sweep: every instance either triggers the fast branch or is refused at the exploration cap, so the answer rate is carried entirely by the fast branch and the refusal decision rather than by the fallback path, but the branch is retained for the logical completeness argued in Section~\ref{sec:method-branches}. 
\subsubsection{The decision map and the tuning of the practical width}\label{sec:experiments-tuning}\label{sec:experiments-phase}

\begin{figure*}[htbp]
  \centering
  \includegraphics[width=\textwidth]{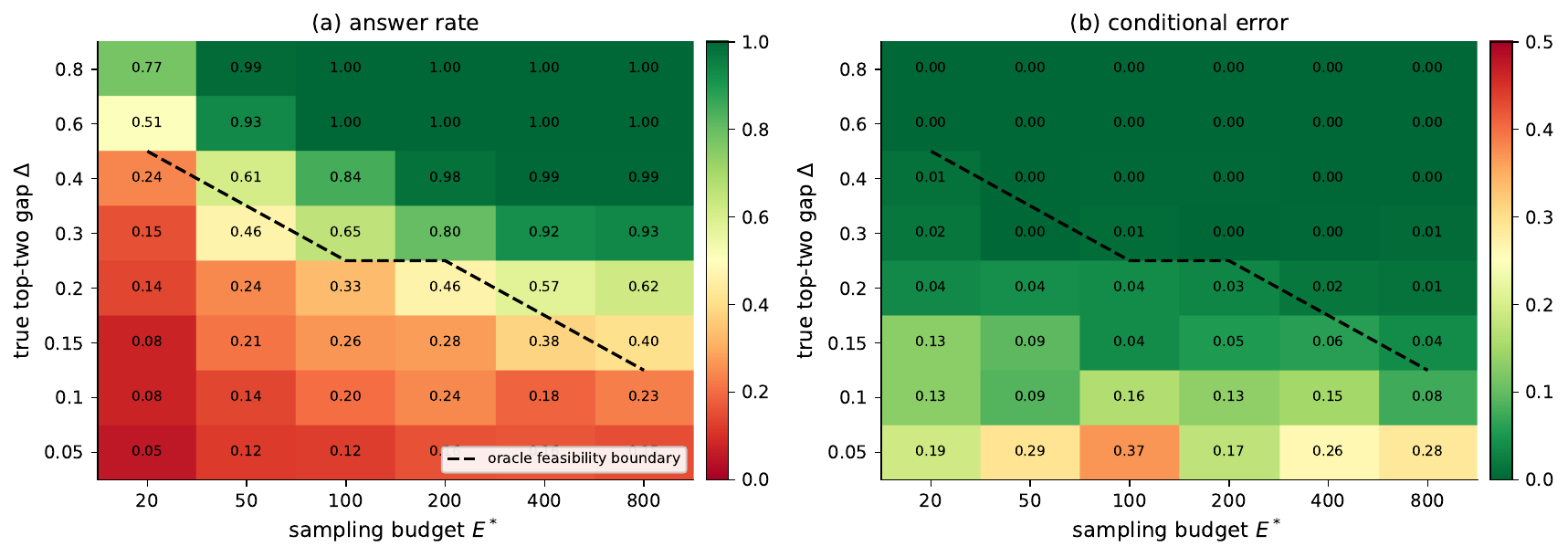}
  \caption{Decision map of the practical mode on the $(\Delta, E^*)$ plane, 400 seeds per cell. Left: answer rate. Right: conditional error on the answered instances (a dash marks a cell in which the rule answered nothing). The dashed contour in both panels is the oracle feasibility boundary of Section~\ref{sec:experiments-design}: cells above and to the right of it are instances a budget-feasible answer exists for.}
  \label{fig:phase}
\end{figure*}

The difficulty sweep varies one axis at a time. 
Crossing it with the budget gives a single decision map, and Figure~\ref{fig:phase} is that map: the answer rate and the conditional error on the $(\Delta, E^*)$ plane, with the oracle feasibility criterion of Section~\ref{sec:experiments-design} overlaid as a dashed contour. 

Three features of the map matter. 
First, the answer region is monotone in both axes but with very different slopes: along $\Delta = 0.05$ the answer rate moves only from $0.05$ to $0.15$ when the budget grows 40-times, whereas along $\Delta = 0.4$ it moves from $0.24$ to $0.99$, because on the first row no budget in the swept range makes the instance feasible while on the second the boundary sits at $E^* \approx 50$. 
Second, the rule's answer region has the same shape as the oracle's feasible region but not the same boundary: on the infeasible rows the rule still answers $5$--$24\%$ of the instances, and on the feasible rows it still refuses part of them, so the two boundaries are within one grid step of each other without coinciding. 
Third, the errors are concentrated where the rule answers rarely: inside the feasible region the conditional error is at most $1.6\%$ for $\Delta \ge 0.3$ and exactly zero for $\Delta \ge 0.6$, whereas at $\Delta = 0.05$, where the rule answers only a small fraction, the answers it does give carry errors of $19$--$37\%$. 
That asymmetry is the whole argument in one panel: the rule answers most where it is right and least where it is wrong. 

\begin{figure*}[htbp]
  \centering
  \includegraphics[width=\textwidth]{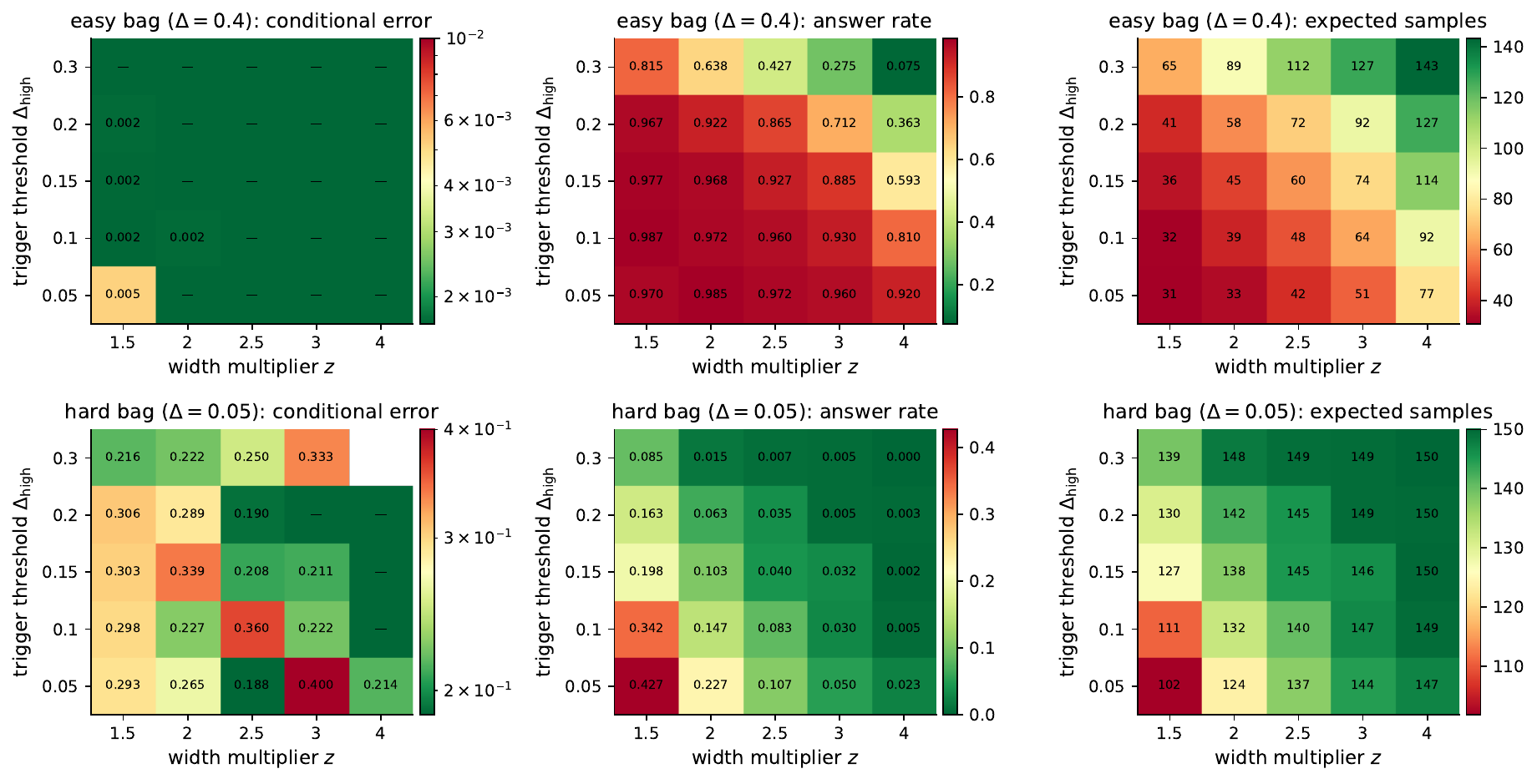}
  \caption{Parameter-tuning heatmaps for the practical width mode, three panels per row (conditional error, answer rate, expected sample count). Top: an easier instance with $\Delta=0.4$. Bottom: a harder instance with $\Delta=0.05$. Axes are the width multiplier $z$ and the trigger threshold $\Delta_{\mathrm{high}}$; 600 seeds per cell. Cells in which fewer than five instances were answered are drawn but their conditional error is not interpretable.}
  \label{fig:heatmap}
\end{figure*}

Tuning the practical mode trades answer rate against mis-trigger control, and the trade-off is visible directly in the $z \times \Delta_{\mathrm{high}}$ grid of Figure~\ref{fig:heatmap}, whose numbers are listed in Table~\ref{tab:tuning}: on the easier instance the conditional error stays below $0.5\%$ everywhere on the grid, and the answer rate falls monotonically as either $z$ or $\Delta_{\mathrm{high}}$ grows, from $98.5\%$ at $(z,\Delta_{\mathrm{high}}) = (2,0.05)$ to $7.5\%$ at $(4,0.3)$. 

On the harder instance the same trade-off is the only thing the grid records, because the rule refuses most instances throughout: the answer rate falls from $42.7\%$ at $z = 1.5$ to below $1\%$ at $z \ge 3$, and the conditional error of the few answered cells is not stable enough to be read as a trend. 
We therefore recommend $\Delta_{\mathrm{high}} \in [0.1,0.2]$ with $z = 2$ for interactive deployment and $z \ge 3$ when stricter mis-trigger control is required, which is the same recommendation as in Section~\ref{sec:method-width}. 
\begin{table*}[htbp]
\caption{Parameter-tuning grid for the practical width, 600 seeds per cell. ``easy'' is $\Delta=0.4$ and ``hard'' is $\Delta=0.05$; the conditional error in cells with few answered instances is not interpretable and is reported only for completeness.}
\label{tab:tuning}
\small
\centering
\setlength{\tabcolsep}{4pt}
\begin{tabular}{@{}cc|ccc|ccc@{}}
\toprule
\multirow{2}{*}{$z$} & \multirow{2}{*}{$\Delta_{\mathrm{high}}$}
& \multicolumn{3}{c|}{easy ($\Delta=0.4$)}
& \multicolumn{3}{c}{hard ($\Delta=0.05$)} \\
\cmidrule(lr){3-5}\cmidrule(lr){6-8}
& & Cond.\ err. & Answer & Mean $n$ & Cond.\ err. & Answer & Mean $n$ \\
\midrule
1.5 & 0.05 & 0.0052 & 0.970 & 30.8  & 0.2930 & 0.427 & 101.8 \\
1.5 & 0.1  & 0.0017 & 0.987 & 31.6  & 0.2976 & 0.342 & 110.8 \\
1.5 & 0.15 & 0.0017 & 0.977 & 35.9  & 0.3025 & 0.198 & 126.6 \\
1.5 & 0.2  & 0.0017 & 0.967 & 40.9  & 0.3061 & 0.163 & 130.5 \\
1.5 & 0.3  & 0.0000 & 0.815 & 64.8  & 0.2157 & 0.085 & 139.0 \\
2   & 0.05 & 0.0000 & 0.985 & 33.5  & 0.2647 & 0.227 & 123.8 \\
2   & 0.1  & 0.0017 & 0.972 & 39.3  & 0.2273 & 0.147 & 132.3 \\
2   & 0.15 & 0.0000 & 0.968 & 45.3  & 0.3387 & 0.103 & 137.9 \\
2   & 0.2  & 0.0000 & 0.922 & 58.5  & 0.2895 & 0.063 & 142.1 \\
2   & 0.3  & 0.0000 & 0.638 & 89.1  & 0.2222 & 0.015 & 148.1 \\
2.5 & 0.05 & 0.0000 & 0.972 & 41.7  & 0.1875 & 0.107 & 137.4 \\
2.5 & 0.1  & 0.0000 & 0.960 & 48.5  & 0.3600 & 0.083 & 140.3 \\
2.5 & 0.15 & 0.0000 & 0.927 & 60.4  & 0.2083 & 0.040 & 145.1 \\
2.5 & 0.2  & 0.0000 & 0.865 & 72.0  & 0.1905 & 0.035 & 145.5 \\
2.5 & 0.3  & 0.0000 & 0.427 & 112.2 & 0.2500 & 0.007 & 149.1 \\
3   & 0.05 & 0.0000 & 0.960 & 51.1  & 0.4000 & 0.050 & 144.4 \\
3   & 0.1  & 0.0000 & 0.930 & 63.5  & 0.2222 & 0.030 & 146.5 \\
3   & 0.15 & 0.0000 & 0.885 & 74.3  & 0.2105 & 0.032 & 146.0 \\
3   & 0.2  & 0.0000 & 0.712 & 92.4  & 0.0000 & 0.005 & 149.4 \\
3   & 0.3  & 0.0000 & 0.275 & 127.3 & 0.3333 & 0.005 & 149.3 \\
4   & 0.05 & 0.0000 & 0.920 & 77.2  & 0.2143 & 0.023 & 147.2 \\
4   & 0.1  & 0.0000 & 0.810 & 92.2  & 0.0000 & 0.005 & 149.4 \\
4   & 0.15 & 0.0000 & 0.593 & 114.0 & 0.0000 & 0.002 & 149.8 \\
4   & 0.2  & 0.0000 & 0.363 & 126.5 & 0.0000 & 0.003 & 149.6 \\
4   & 0.3  & 0.0000 & 0.075 & 143.3 & ---    & 0.000 & 150.0 \\
\bottomrule
\end{tabular}
\end{table*}

The one reading of the practical mode that moves against the rule is the mis-refusal rate at the largest budget: at $\Delta = 0.2$ and $E^* = 500$, by which point the instance is feasible, the rule answers $57.1\%$ of the instances with a $1.1\%$ conditional error and still mis-refuses $42.9\%$ of the feasible ones. 
The practical width is the cause, not the criterion: the criterion is evaluated at $d_{\mathrm{use}} = \max(0, \hat\Delta_\tau - w_\tau)$ with the $z = 2$ heuristic width, so at moderate budgets the endpoint is conservative enough to move a borderline instance across the boundary. 
We report this as a measured cost of the practical mode rather than as a defect of the criterion, and raising $z$ is what trades it back against the answer rate, as the grid above shows. 

The chooser is the second tuning axis, and it is the one with the larger effect. 
The pessimistic chooser is not usable at these budgets: it produces a formally valid certificate but judges almost every instance infeasible, collapsing the answer rate, whereas the optimistic chooser reaches a $97.2\%$ answer rate with a $0.17\%$ conditional error and $39.3$ samples on average, $2.4$ times the oracle cost. 
This is the empirical form of the trade-off stated in Section~\ref{sec:theory}: the pessimistic chooser is the configuration the theorems cover, and the optimistic chooser is the configuration that runs. 

\subsubsection{Ablation studies}\label{sec:experiments-cert}

In this part, we try to separate what each configuration contributes in the two experiments following: the first prices the certified mode by running the theorem configuration, and the second prices the fast branch by running it only.

\textbf{The certified mode against the practical mode.}

\begin{figure*}[htbp]
  \centering
  \includegraphics[width=\textwidth]{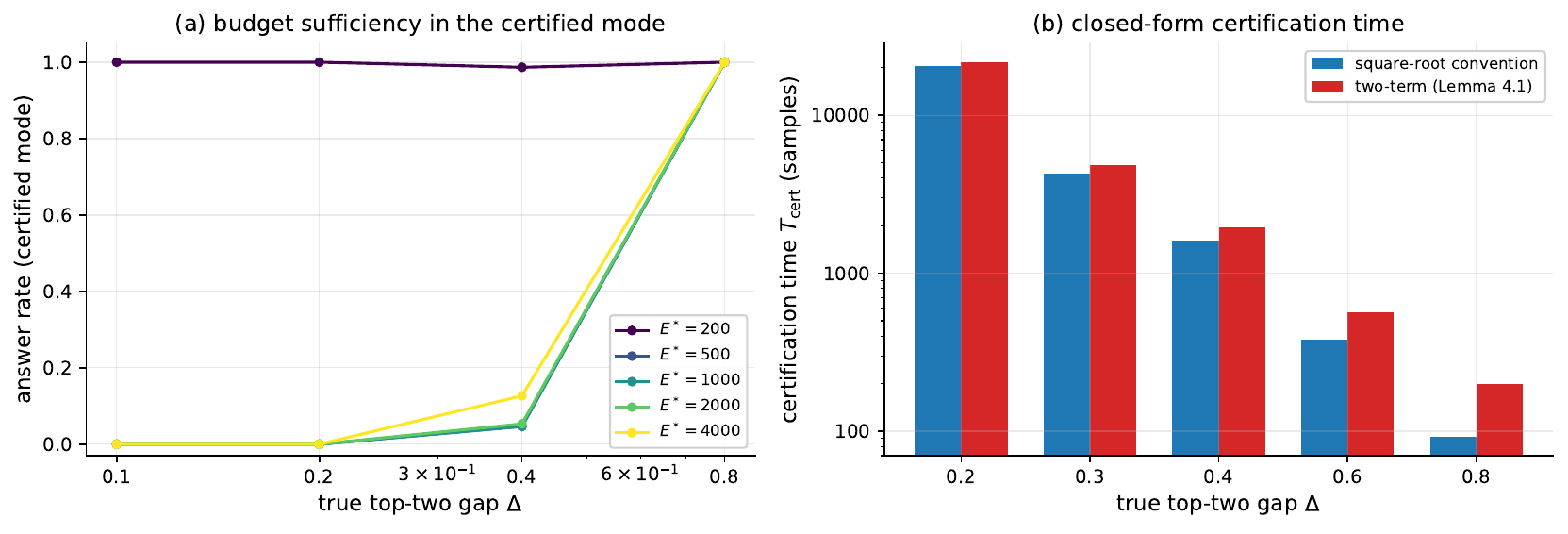}
  \caption{Budget sufficiency in the certified mode. Left: at a given budget the answer rate increases with $\Delta$; no incorrect answer was observed in these runs, so no conditional-error panel is drawn. Right: the theoretical certification time $T_{\mathrm{cert}}$, computed from $w_t \le \Delta - \Delta_{\mathrm{high}}$ under the two conventions of Lemma~\ref{lem:cs}.}
  \label{fig:adequacy}
\end{figure*}

Figure~\ref{fig:adequacy} examines the certified mode directly, and Table~\ref{tab:cert-times} gives the underlying closed-form inputs. 

The certified mode answers only instances whose gap is large enough to certify within the budget: at $E^* = 1000$ the answer rate is zero for $\Delta \le 0.2$, $5\%$ at $\Delta = 0.4$ and $100\%$ at $\Delta = 0.8$, and raising the budget to $E^* = 4000$ moves the $\Delta = 0.4$ point only to $13\%$. 
No incorrect answer was observed in any certified cell, which is the qualitative content of Corollary~\ref{cor:trigger} on instances whose certified trigger fires. 
The asymmetry is the point: the certified width is not a heuristic that can be tightened by spending more samples on a hard instance, because the certification cost grows as $(\Delta-\Delta_{\mathrm{high}})^{-2}$ and the instance is simply not certifiable at these budgets. 

Solving $w_t \le \hat\Delta_t - \Delta_{\mathrm{high}}$ with the certified width yields the smallest $t$ at which the trigger can fire, that is $T_{\mathrm{cert}}$, and with the two-point-family variance $\sigma^2 = 1-\Delta^2$ the effective gap is $\Delta - \Delta_{\mathrm{high}}$ rather than $\Delta$, which is why the scaling is not exactly $\Delta^{-2}$. 
Concretely, the square-root convention gives $2.05\times10^{4}$, $1.60\times10^{3}$ and $9.20\times10^{1}$ samples for $\Delta = 0.2$, $0.4$ and $0.8$, and retaining the additive term of Lemma~\ref{lem:cs} raises these to $2.18\times10^{4}$, $1.96\times10^{3}$ and $1.99\times10^{2}$. 
Both sets of values quantify the same tension: a certified answer on a moderately hard instance costs two to three orders of magnitude more samples than a budgeted one, and the gap is a property of the guarantee rather than of the implementation. 
Finally, the search for a window in the fast branch has to be distinguished from the trigger: the trigger can fire only after $T_{\mathrm{cert}}$, and the dynamic program is what decides whether a window then exists. 
We report both conventions because the choice is a modeling decision about which form of the empirical-Bernstein bound is used, not a numerical detail: the ratio between the two grows monotonically with the gap, from $1.06$ at $\Delta = 0.2$ to $2.16$ at $\Delta = 0.8$, while the scaling in $(\Delta-\Delta_{\mathrm{high}})^{-2}$ is common to both, and it is under the square-root-only convention that a constant batch of draws would certify a gap with zero width. 
\begin{table*}[htbp]
\caption{Certification time $T_{\mathrm{cert}}$ under the square-root convention used by the implementation and the figures, and under the two-term convention of Lemma~\ref{lem:cs}, with $\delta=0.05$, $\Delta_{\mathrm{high}}=0.1$, $\bar K=10$ and $\sigma^2=1-\Delta^2$. All values are closed-form and follow from Section~\ref{sec:method-width}; no experiment is involved, so there is no sampling uncertainty to report.}
\label{tab:cert-times}
\small
\centering
\begin{tabular}{@{}lccc@{}}
\toprule
$\Delta$ & square-root term & two-term (Lemma~\ref{lem:cs}) & ratio \\
\midrule
0.2 & $2.05\times10^{4}$ & $2.18\times10^{4}$ & 1.06 \\
0.3 & $4.27\times10^{3}$ & $4.81\times10^{3}$ & 1.13 \\
0.4 & $1.60\times10^{3}$ & $1.96\times10^{3}$ & 1.22 \\
0.6 & $3.82\times10^{2}$ & $5.68\times10^{2}$ & 1.49 \\
0.8 & $9.20\times10^{1}$ & $1.99\times10^{2}$ & 2.16 \\
\bottomrule
\end{tabular}
\end{table*}

\textbf{The configuration the theorems cover.}\label{sec:experiments-theorem-config}

This second run closes the gap between the runs and the theorems by using the theorem configuration on the same $\Delta$ grid as the practical one, with all four configurations sharing one batch of seeds. 
We use four configurations. 
C1 is the certified width with the pessimistic chooser and the square-root-only convention; C2 is the certified width with the pessimistic chooser and the two-term width of Lemma~\ref{lem:cs}, which is exactly the configuration of the two theorems; C3 is the certified width with the optimistic chooser and the two-term width; and C4 is the practical width ($z=2$) with the optimistic chooser, the configuration already used in Sections~\ref{sec:experiments-sweep} and~\ref{sec:experiments-tuning}. 

\begin{table*}[htbp]
\caption{The four configurations defined in this subsection, on a shared $\Delta$ grid with 300 seeds per cell, $E^*=1000$, $T_{\mathrm{cap}}=250$, $\mathrm{min\_t}=8$, $p^*=0.05$, $\delta=0.05$, $\Delta_{\mathrm{high}}=0.1$ and $K=10$; the four share one seed batch. C1: certified width, pessimistic chooser, square-root-only convention. C2: certified width, pessimistic chooser, two-term width (the configuration the theorems cover). C3: certified width, optimistic chooser, two-term width. C4: practical width ($z=2$), optimistic chooser. ``Deg.\ trig.'' counts the seeds whose trigger fired on a batch with zero plug-in variance.}
\label{tab:theorem-config}
\small
\centering
\begin{tabular}{@{}lccccccc@{}}
\toprule
Config & $\Delta$ & Answer & Refusal & Mean samples & Mean trigger & Cond.\ err. & Deg.\ trig. \\
\midrule
C1 & 0.05 & 0.000 & 1.000 & 250.0 & 250.0 & 0.0000 & 0 \\
C1 & 0.1 & 0.000 & 1.000 & 250.0 & 250.0 & 0.0000 & 0 \\
C1 & 0.2 & 0.013 & 0.987 & 253.5 & 247.4 & 0.0000 & 2 \\
C1 & 0.4 & 0.743 & 0.257 & 784.5 & 156.6 & 0.0000 & 7 \\
C1 & 0.8 & 1.000 & 0.000 & 838.1 & 18.9 & 0.0000 & 45 \\
C2 & 0.05 & 0.000 & 1.000 & 250.0 & 250.0 & 0.0000 & 0 \\
C2 & 0.1 & 0.000 & 1.000 & 250.0 & 250.0 & 0.0000 & 0 \\
C2 & 0.2 & 0.000 & 1.000 & 250.0 & 250.0 & 0.0000 & 0 \\
C2 & 0.4 & 0.033 & 0.967 & 275.0 & 248.3 & 0.0000 & 0 \\
C2 & 0.8 & 1.000 & 0.000 & 997.2 & 85.0 & 0.0000 & 1 \\
C3 & 0.05 & 0.000 & 1.000 & 250.0 & 250.0 & 0.0000 & 0 \\
C3 & 0.1 & 0.000 & 1.000 & 250.0 & 250.0 & 0.0000 & 0 \\
C3 & 0.2 & 0.000 & 1.000 & 250.0 & 250.0 & 0.0000 & 0 \\
C3 & 0.4 & 0.033 & 0.967 & 248.6 & 248.3 & 0.0000 & 0 \\
C3 & 0.8 & 1.000 & 0.000 & 89.8 & 85.0 & 0.0000 & 1 \\
C4 & 0.05 & 0.183 & 0.817 & 215.2 & 210.5 & 0.3091 & 0 \\
C4 & 0.1 & 0.263 & 0.737 & 211.6 & 199.3 & 0.0886 & 0 \\
C4 & 0.2 & 0.733 & 0.267 & 151.7 & 116.5 & 0.0091 & 2 \\
C4 & 0.4 & 1.000 & 0.000 & 40.4 & 27.5 & 0.0000 & 7 \\
C4 & 0.8 & 1.000 & 0.000 & 13.8 & 8.9 & 0.0000 & 45 \\
\bottomrule
\end{tabular}
\end{table*}

Three readings of Table~\ref{tab:theorem-config} matter. 
First, the configuration the theorems cover is feasible: C2 answers every seed at $\Delta=0.8$ with zero conditional error, at about $997$ samples. 
But it refuses almost everything at $\Delta \le 0.4$, where its answer rate is at most $3.3\%$, so the certified width alone already makes the rule unusable at the budgets of Section~\ref{sec:experiments-sweep}. 
Second, what is expensive is the chooser rather than the width: at the same certified width and two-term width, replacing the pessimistic chooser by the optimistic one (C2$\to$C3) lowers the mean sample count at $\Delta=0.8$ from $997.2$ to $89.8$, a factor of about eleven, with the conditional error still zero. 
Third, the additive term is not optional in practice either: the degenerate-trigger counts logged during the sweep are $45$, $7$ and $2$ at $\Delta=0.8$, $0.4$ and $0.2$ for C1, and $1$, $0$ and $0$ for C2, so the additive term essentially clears the early degenerate triggers. 

\textbf{The fast branch in isolation.}\label{sec:experiments-benefit}

\begin{figure*}[htbp]
  \centering
  \includegraphics[width=\textwidth]{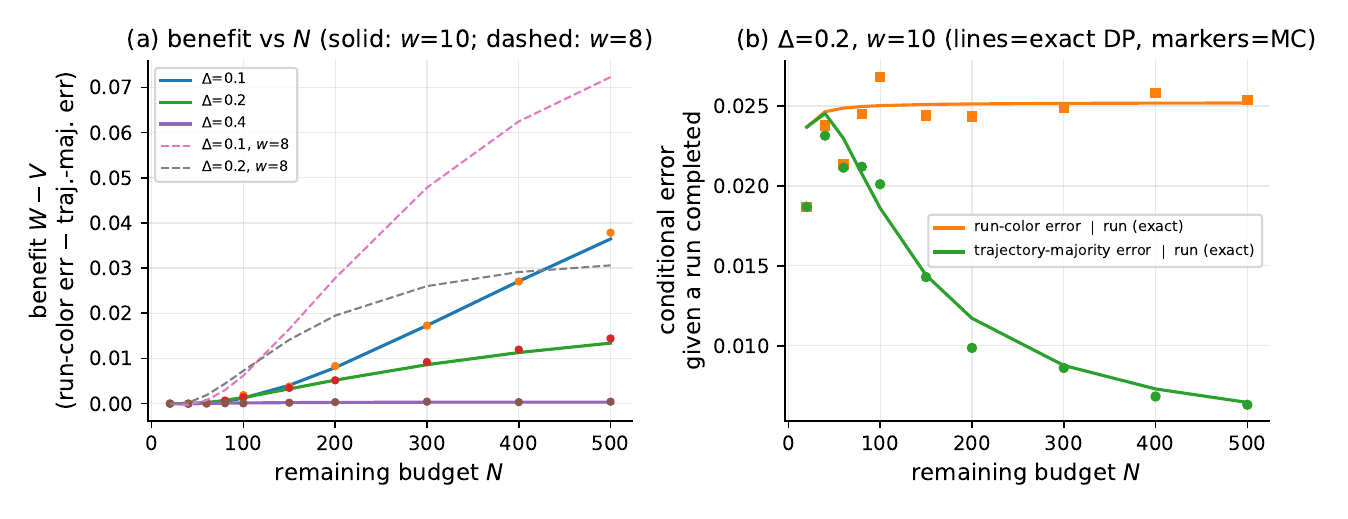}
  \caption{Trajectory-majority benefit in synthetic fast-branch experiments without a language model. Left: the exact gain $W-V$ against the remaining budget $N$ for $\Delta = 0.1, 0.2, 0.4$ and run lengths $w=10$ (solid) and $w=8$ (dashed); markers are Monte Carlo estimates over $40{,}000$ seeds. Right: conditional error rates of the run-color and trajectory-majority rules, given that a run completes ($\Delta=0.2$, $w=10$); lines are exact chain values and markers are Monte Carlo estimates.}
  \label{fig:trajmaj}
\end{figure*}

Proposition~\ref{prp:benefit} attributes the gain of answering with the trajectory majority, rather than with the color that completed the run, to the chain mass $W-V$; we verify the identity in a controlled version of the fast branch that involves no language model, and --- because the same experiment compares BARS against the rule it would be if the run color were the answer --- it also isolates the contribution of the answering rule from the contribution of the stopping rule. 
Figure~\ref{fig:trajmaj} reports the result. 

The gain grows with the remaining budget $N$: at $\Delta = 0.2$ with the dynamic program's typical run length $w=10$ it rises from $1.3\times10^{-3}$ at $N=100$ to $5.2\times10^{-3}$ at $N=200$ and $1.3\times10^{-2}$ at $N=500$, and it was non-negative at every point of the tested grid, which is what Proposition~\ref{prp:benefit} allows for without proving it for every distribution. 
Conditional on a run completing, the trajectory majority halves the error of the run-color rule: at $\Delta=0.2$, $w=10$ and $N=200$ the conditional error falls from $2.5\%$ to $1.2\%$, and the improvement is the mass of the paths on which a minority completes the run while the majority still leads, an event that becomes more likely as the lead is established over more samples. 
The same sweep provides a numerical check on Lemma~\ref{lem:dom}: folding random distributions from the confidence set into a single filler color never produced a candidate that outperformed the folded pessimistic bag, with a maximum deficit below $10^{-5}$ over the tested grid. 
\subsection{Real-dataset experiments}\label{sec:experiments-llm}

Section~\ref{sec:experiments-synth} measured quantities that only a known distribution supplies: the true gap, the oracle feasibility boundary and the mis-refusal rate, none of which can be computed on real questions.
This subsection measures what only a real model supplies.
The pool is produced by \texttt{Qwen2.5-7B} on GSM8K and MATH, sixteen decodes per question and eighty questions per dataset, and every rule listed in Section~\ref{sec:experiments-design} consumes the same pool.
Two quantities change meaning across this boundary, and both are stated where they change: accuracy is now scored against a reference rather than against a known majority, and the mis-refusal rate is not reported at all, because its definition refers to the true gap; the refusals are instead priced ex post, by counting how many withheld questions the pool majority would have answered correctly.

\subsubsection{Setup and parsing}\label{sec:experiments-llm-setup}

In this part, each response is parsed by a priority-based extraction rule and the extracted final answer is normalized to a color key. 
MATH answers receive light normalization, and the few malformed multi-candidate responses are counted as errors rather than discarded, so that the error figures are not flattered by parsing. 
The mis-refusal rate cannot be computed on real questions, because its definition refers to the true gap (Section~\ref{sec:experiments-design}), so we report no value for it here and instead price the refusal ex post in the results below. 
This section runs under its own parameter set, namely $E^*=16$, $p^*=0.02$, $\Delta_{\mathrm{high}}=0.3$, $\delta=0.05$ and the practical width ($z=2$), so the certified-mode level $\delta$ is not a gate on the running system here, and the synthetic defaults of Section~\ref{sec:experiments-synth} do not transfer to these runs. 

\subsubsection{Results}\label{sec:experiments-llm-results}

\begin{figure*}[htbp]
  \centering
  \includegraphics[width=\textwidth]{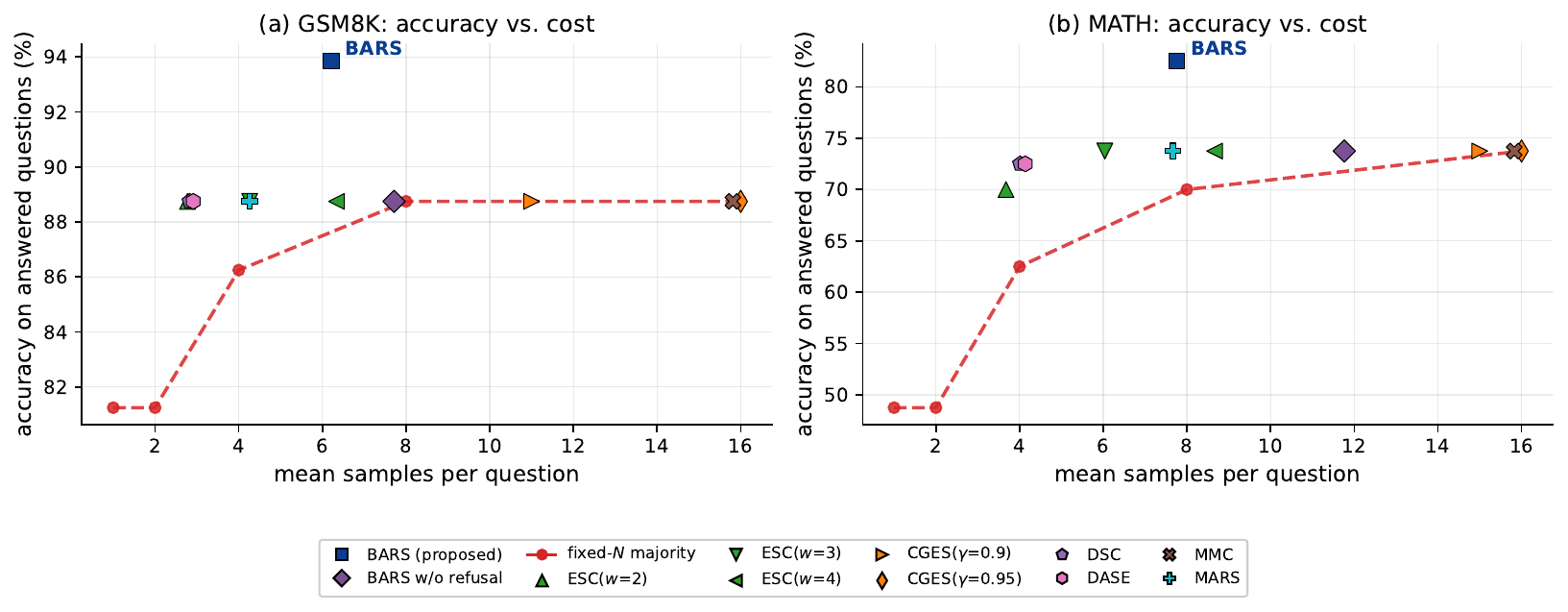}
  \caption{Real-LLM results on GSM8K and MATH in the practical mode, with a pre-generated pool of $E^*=16$ decodes per question and 80 questions per dataset. (a) GSM8K and (b) MATH: accuracy on the answered questions against the mean number of samples, one marker per rule, with the fixed-$N$ majority joined by a dashed line as a reference and the rule under test drawn as a filled square. 
  }
  \label{fig:llm}
\end{figure*}

Figure~\ref{fig:llm} and Table~\ref{tab:llm} report the results. 

On GSM8K the rule answers $81.2\%$ of the questions after 6.21 samples on average, with an accuracy of $93.8\%$ on those answers, against $88.8\%$ for the fixed-16 majority from the same pool. 
Because the two rules do not spend the same amount, the matched-cost comparison is the informative one: at 6.21 samples the closest fixed baseline is $N=8$, which reaches $88.8\%$, so at the same average cost the rule buys $5.1$ points of accuracy on the answered questions, and the currency it pays in is coverage. 
Over the dataset as a whole --- the denominator that includes the refused questions --- the rule draws $22.3\%$ fewer samples than fixed-8. 
On MATH the same comparison is sharper: the rule answers $50.0\%$ of the questions at 7.76 samples with $82.5\%$, where the matched-cost baseline $N=8$ reaches $70.0\%$ and the full-budget $N=16$ baseline reaches $73.8\%$ over the whole set. 
The full-coverage baselines report those numbers only because they answer everything, so their accuracy is a ceiling rather than a choice. 
The refusal is not free, and the pool lets us price it ex post rather than only in principle: of the 15 questions the rule refuses on GSM8K, 10 would have been answered correctly by the pool majority, and of the 40 it refuses on MATH, 26 would have been. 
So the rule gives up 18.8\% of the GSM8K set and 50.0\% of the MATH set, and we report that as the measured price of the output channel rather than as a success of it; the Wilson intervals of the rule and of the full-coverage baselines overlap on both datasets, so the accuracy gain is not statistically established at these sample sizes. 
Note that this ex post count is not the mis-refusal rate defined in Section~\ref{sec:experiments-design}: the latter is stated against the analyst-level feasibility event $\mathcal{A}$, which is built from the true gap and therefore cannot be evaluated on real questions, whereas the count here only says that the answers the rule withheld would in hindsight have been right. 
Scoring each refusal as an error rather than as a withheld answer places the rule at $76.2\%$ on GSM8K and $41.2\%$ on MATH over the whole set, below every full-coverage baseline; we report the figure because it is the same trade-off seen from the other side, and a deployment that must answer every question should not use this rule. 
Finally, the competitor rows carry the same reading as the synthetic ones: the early-stopping rules match the full-budget accuracy at a fraction of the cost but cannot exceed it, because exceeding it would require declining some questions; on MATH, where the pool is genuinely fragmented, that ceiling is what separates them from the rule under test. 
The two datasets therefore differ in how much the refusal costs rather than in whether the rule behaves the same way, and we read that as the cross-domain content of the finding rather than as a per-dataset tuning artifact. 

Reading the competitor rows rule by rule, each of them lands at or below the full-budget accuracy, because none of them can decline a question: 
The distinction is not that BARS is cheaper on every cell but that its cheapness and its accuracy are produced by one decision: on MATH the competitors spend their budget on questions whose answer distribution they cannot resolve, whereas BARS declines them and spends its budget on the rest. 

\begin{table*}[htbp]
\caption{Real-LLM results, \texttt{Qwen2.5-7B}, 80 questions per dataset, pre-generated pool $E^*=16$, all rules on the same pool. ``ans.'' is the answer rate, ``acc.'' the accuracy on the answered questions, and $n$ the mean sample count over all questions of the dataset, refused questions included. Accuracy intervals are Wilson $95\%$ intervals and are reported in the text.}
\label{tab:llm}
\small
\centering
\begin{tabular}{@{}lcccccc@{}}
\toprule
Rule & \multicolumn{3}{c}{GSM8K} & \multicolumn{3}{c}{MATH} \\
\cmidrule(lr){2-4}\cmidrule(lr){5-7}
 & ans.\ (\%) & acc.\ (\%) & $n$ & ans.\ (\%) & acc.\ (\%) & $n$ \\
\midrule
Fixed-1 & 100.0 & 81.2 & 1.00 & 100.0 & 48.8 & 1.00 \\
Fixed-2 & 100.0 & 81.2 & 2.00 & 100.0 & 48.8 & 2.00 \\
Fixed-4 & 100.0 & 86.2 & 4.00 & 100.0 & 62.5 & 4.00 \\
Fixed-8 & 100.0 & 88.8 & 8.00 & 100.0 & 70.0 & 8.00 \\
Fixed-16 & 100.0 & 88.8 & 16.00 & 100.0 & 73.8 & 16.00 \\
ESC($w$=2) & 100.0 & 88.8 & 2.77 & 100.0 & 70.0 & 3.67 \\
ESC($w$=3) & 100.0 & 88.8 & 4.26 & 100.0 & 73.8 & 6.04 \\
ESC($w$=4) & 100.0 & 88.8 & 6.34 & 100.0 & 73.8 & 8.66 \\
CGES($\gamma$=0.9) & 100.0 & 88.8 & 11.00 & 100.0 & 73.8 & 15.00 \\
CGES($\gamma$=0.95) & 100.0 & 88.8 & 16.00 & 100.0 & 73.8 & 16.00 \\
DSC & 100.0 & 88.8 & 2.83 & 100.0 & 72.5 & 4.01 \\
DASE & 100.0 & 88.8 & 2.91 & 100.0 & 72.5 & 4.14 \\
MMC & 100.0 & 88.8 & 15.81 & 100.0 & 73.8 & 15.82 \\
MARS & 100.0 & 88.8 & 4.26 & 100.0 & 73.8 & 7.66 \\
BARS-NoRefusal & 100.0 & 88.8 & 7.71 & 100.0 & 73.8 & 11.76 \\
BARS & 81.2 & 93.8 & 6.21 & 50.0 & 82.5 & 7.76 \\
\bottomrule
\end{tabular}
\end{table*}

\subsection{Discussion}\label{sec:experiments-discussion}



The synthetic and the deployed runs exhibit the same qualitative behavior: easy instances are answered quickly and hard ones are refused. 
The competitor comparison adds three findings that the internal audit alone could not produce. 
First, where no answer is budget-feasible the advantage of BARS is not a matter of degree but of kind: on the smallest gap of the sweep, which the oracle calls infeasible, the rule refuses $87\%$ of the instances, whereas the five fixed-budget rules assert an answer that is wrong in $33$--$44\%$ of them, and no parameter of theirs could have produced the honest output because none of them has a third output to produce it with. 
Second, on instances that are answerable, the saving comes from stopping rather than from answering: BARS matches the conditional error of the fixed-budget baseline and of LUCB on the easy cells at $14$--$39$ samples against their $97$--$200$, and the difference is entirely in how long the rules wait rather than in what they answer. 
Third, the elimination-based BAI rules are the wrong tool for this problem rather than a weak version of the right one: on the hard cells of the sweep their error ($35$--$44\%$ at $\Delta \le 0.05$) exceeds that of the fixed-budget majority ($21$--$33\%$), because a fixed-budget arm-identification rule must spend its budget eliminating arms it cannot separate and then recommend one of them. 

The fixed-budget baseline cannot recover hard instances even when it spends the whole budget, whereas the adaptive rule gives up on a subset of questions in exchange for higher accuracy on the rest. 

\section{Conclusion and future work}\label{sec:conclusion}

This paper formalizes multiple-sampling majority voting as majority identification under an unknown distribution with a sampling budget and a target error rate, and proposes \textbf{BARS} (\textbf{B}udget-\textbf{A}ware \textbf{R}efusal \textbf{S}ampling), a sequential rule driven by the top-two gap $\Delta$ and its plug-in variance. 
The rule returns one of three outputs --- an early stop, a majority vote after the budget is spent, or an explicit refusal --- and it can be run in two configurations: the certified mode, for which we prove bounds on the joint answering error and on the mis-refusal probability, and the practical mode, which is empirically calibrated and for which no theorem is claimed. 
Each certified-mode guarantee is backed by a proof; the practical mode is backed by measurement, and the distance between the two is reported by this paper. 

The experiments divide into two settings and one discussion, and the findings follow that division. 
First, the synthetic sweeps over random seeds show the difficulty-driven behavior that the certified-mode analysis predicts: large gaps are settled far below the budget, small gaps are refused, and the boundary moves with the $(\Delta-\Deltahigh)^{-2}$ certification cost. 
Second, certification itself is expensive at these problem sizes: a certified answer costs two to three orders of magnitude more samples than a budgeted one, which is why the certified mode refuses at small budgets instead of answering badly, and why the two-point lower bound is a statement about budgets rather than about algorithms. 
Third, on a real deployment budget the rule behaves as the abstraction predicts: on GSM8K it draws $22.3\%$ fewer samples over the pool than fixed-8 sampling (denominator: all 80 questions, refused questions included) at the cost of declining $18.8\%$ of the questions, and on MATH it refuses half of the questions while lifting the answered accuracy from $70.0\%$ to $82.5\%$, so that refusal converts questions the configuration cannot handle from silent errors into an actionable output. 

The two theorems' own configuration is also measured now: at $\Delta=0.8$ the certified width with the pessimistic chooser is feasible but spends essentially the whole sampling budget, and changing only the chooser to the optimistic one reduces the mean sample count to $89.8$ (Section~\ref{sec:experiments-theorem-config}). 

Five directions follow from the limitations we report, ordered by how much of the remaining uncertainty each would remove. 
First, the certified mode needs tighter time-uniform widths, for instance self-normalised or betting-based confidence sequences \citep{howard2021,waudbysmith2023}, since its current constant is what rules it out at deployment-scale budgets. 
Second, an independent reference direction, generated separately from the rollouts the rule consumes, would remove the reference dependence of every accuracy figure reported here. 
Third, a low-repetition or multi-peaked answer pool would test the abstraction where it is weakest, because the current pool's concentration is partly repetition rather than semantic certainty. 
Fourth, decoupling $\Delta$ from answer quality, for instance by weighting answers the model judges plausible, would address the structural boundary that no bound in this paper can detect: a model that is confidently and consistently wrong. 
Fifth, online batching would use the stopping-time distribution as a scheduling signal when many questions share a GPU, and larger models and budgets would test whether the deployed behavior we report survives at scale. 

\section*{CRediT authorship contribution statement}
\textbf{Shenxi Liu}: Conceptualization, Methodology, Software, Validation, Formal analysis, Writing -- original draft. \textbf{Yiwei Li}: Methodology, Validation, Writing -- review \& editing. \textbf{Mingyang Zhao}: Software, Data curation, Visualization, Writing -- review \& editing. \textbf{Yuhang Tian}: Investigation, Writing -- review \& editing. \textbf{Kan Li}: Supervision, Funding acquisition, Project administration, Writing -- review \& editing.

\section*{Declaration of competing interest}
The authors declare that they have no known competing financial interests or personal relationships that could have appeared to influence the work reported in this paper.

\section*{Acknowledgements}
This research was supported by the Beijing Natural Science Foundation (No. 4262065), and the National Natural Science Fundation of China (No. 62676029, 62171183).

\section*{Data availability}
\begin{sloppypar}
All experimental code related to this study can be accessed at \url{https://github.com/LtColLiuPeiqiang/adaptive-stopping/}.
\end{sloppypar}

\bibliographystyle{elsarticle-num}
\bibliography{reference}

\appendix
\appendix
\section{Proofs}\label{app:proofs}

The proofs follow the order in which the results are used, so that no proof relies on a statement proved after it. 
Throughout, $\Gcal$ denotes the good event of Lemma~\ref{lem:cs} and Corollary~\ref{cor:trigger}, and constants absorbed into the width are flagged where they are used. 

\subsection{Proof of the confidence sequence with the additive term (Lemma~\ref{lem:cs})}\label{app:proof-cs}

At each time $t$, the empirical-Bernstein inequality of \cite{maurer2009} (Theorem~4, for variables bounded in $[-1,1]$) gives, for any $c_t>0$,
\[
\Prob\Big(|\bar X_t - \mu| \;\ge\; \sqrt{\frac{2\hat\sigma_t^2\,c_t}{t}} + \frac{7c_t}{3t}\Big) \;\le\; 2e^{-c_t}.
\]
Substituting $c_t = 2\ln t + \ln\frac{\pi^2}{3\delta}$ turns the right-hand side into
\[
2e^{-2\ln t}\cdot e^{-\ln\frac{\pi^2}{3\delta}} = \frac{6\delta}{\pi^2}\cdot\frac{1}{t^2},
\]
and a union bound over $t \ge 2$, using $\sum_{t\ge 1} t^{-2} = \pi^2/6$, yields $\Prob(\exists\, t\ge 2: |\bar X_t - \mu| \ge w_t) \le \delta$. 
This is the mechanism of time-uniformity: the per-test mass $\frac{6\delta}{\pi^2 t^2}$ is a Cauchy sequence in $t$, so the total over all times is exactly the nominal level. 
Both terms of $w_t$ are needed: the square-root term shrinks with the plug-in variance and would vanish on a constant batch of draws, whereas the additive term $7c_t/(3t)$ is a strictly decreasing positive function of $t$ and keeps the bound informative at every time (Remark~\ref{rem:floor}). 

\subsection{Proof of the top-two gap control and of the certified trigger (Corollary~\ref{cor:trigger})}\label{app:proof-order}

The order statistics $p_{(k)}$ are $1$-Lipschitz in the $\ell_\infty$ sense: for any pair of distributions $p, \hat p$ and any rank $k$,
\[
\big|\hat p_{(k)} - p_{(k)}\big| \le \max_c |\hat p_c - p_c|.
\]
Applying this at ranks $k=1$ and $k=2$ and using the triangle inequality gives $|\hat\Delta_t - \Delta| \le 2\max_c|\hat p_c - p_c|$. 
To transfer Lemma~\ref{lem:cs} to the colors, a union bound over the at most $\bar K$ of them allocates the per-test mass $6\delta/(\bar K\pi^2 t^2)$ to each color, so the joint event $\{\exists\, t\ge 2, c:\ |\hat p_{t,c} - p_c| \ge w_t^0\}$ has probability at most $\delta$, where $w_t^0$ is the per-color half-width of Lemma~\ref{lem:cs} with $\bar K$ inserted into $c_t$. 
Multiplying $w_t^0$ by the Lipschitz factor $2$ of the top-two gap gives the width of Corollary~\ref{cor:trigger}, and on the resulting event $\Gcal$ every trigger $\hat\Delta_t - w_t \ge \Deltahigh$ certifies $\Delta \ge \Deltahigh$. 

\subsection{Proof of the pessimistic dominance (Lemma~\ref{lem:dom})}\label{app:proof-dom}

We split the argument into a monotonicity step over the two-color family and a folding step. 

\textbf{Step 1 (two-color monotonicity).} For a two-color bag with majority probability $m$, the fast branch is a Markov chain on the state $(r_1, r_2, d)$, where $r_1, r_2 \in \{0, \dots, w-1\}$ are the current run lengths of the majority color and the filler and $d$ is the count gap $N_1 - N_2$; the chain is killed when either run length reaches $w$ or $N$ steps elapse, and the payoff is ``error $= d \le 0$ at the kill''. 
Order the states by $(d \uparrow,\ r_1 \uparrow,\ r_2 \downarrow)$. 
Each transition is monotone in this order: a majority draw maps $(d, r_1, r_2)$ to $(d+1, r_1+1, 0)$ or kills with payoff $d+1 \le 0$, whereas a filler draw maps to $(d-1, 0, r_2+1)$ or kills with payoff $d-1 \le 0$. 
Hence, by the standard comparison theorem for monotone Markov chains (equivalently, by backward induction on the remaining horizon: the value function is non-decreasing in the order, and increasing $m$ shifts probability mass from filler draws to majority draws, which only raises the value), the error probability is non-increasing in $m$. 
The hardest two-color bag with $p_{(1)} \ge m$ is therefore the one whose majority probability equals $m$, namely $q^{\mathrm{pess}}$. 

\textbf{Step 2 (folding).} Fix a majority probability $m' \ge m$ and couple any distribution $p$ with $p_{(1)} = m'$ to the folded two-color bag $q_{m'}$ on a common uniform sequence: every minority outcome of $p$ maps to the single filler color of $q_{m'}$. 
At every time $t$ the majority count and the total minority count coincide in both processes, so the gap process $d_t = N_{(1)}(t) - (\text{total minority count})(t)$ is identical; the majority run lengths coincide, and the folded filler run length is at least every individual minority run length. 
Consequently the unfolded error event at a fixed time, $\{\exists c' \ge 2: N_{c'}(t) \ge N_{(1)}(t)\}$, is contained in the folded error event $\{d_t \le 0\}$. 
The only remaining difference is the killing time: the folded process is killed by a filler $w$-run no later than the unfolded one. 
On the paths where the folded process is killed by a majority run the two processes agree exactly; on the paths where it is killed by a filler run while the majority still leads ($d > 0$), the unfolded process continues, and the folded error event no longer witnesses the unfolded one. 
On those paths the unfolded process can err only if some minority overtakes the majority against a negative drift $-\Delta_c \le -\Delta$. 
The overshoot time $\tau = \min\{t \ge \tau_{\mathrm{fold}}: \exists c' \ge 2,\ N_{c'}(t) \ge N_{(1)}(t)\}$ is an upcrossing of the negative-drift random walk $S_t^{(c')} = \sum_{i=\tau_{\mathrm{fold}}+1}^{t} (\mathbf 1\{X_i = c'\} - \mathbf 1\{X_i = 1\})$; by Hoeffding's inequality on its increments (bounded in $[-1,1]$, drift $-\Delta_c$) combined with the reflection principle, $\Prob(\tau \le N \mid \Gcal) \le 2 e^{-N\Delta^2/2}$ for each minority color, and a union over the at most $\bar K - 1$ minority colors gives $2(\bar K-1)e^{-N\Delta^2/2} \le (\bar K-1)e^{-N d_{\mathrm{use}}^2/8} = \rho_N$, where the last step uses $2e^{-N\Delta^2/2} \le e^{-N\Delta^2/8}$ and $d_{\mathrm{use}} \le \Delta$ on $\Gcal$. 
Combining with Step~1 gives the statement of the lemma. 

\subsection{Proof of the fast-branch error (Theorem~\ref{thm:fast})}\label{app:proof-fast}

The error event at the branch stop splits into three disjoint events (Section~\ref{sec:method-branches}): (\textit{a})~the first $w$-run is completed by a minority color and the minority is the trajectory leader at the stop; (\textit{b})~the first $w$-run is completed by the majority color yet the majority is not the trajectory leader at the stop; and (\textit{c})~no color completes a $w$-run within the remaining $N$ samples, in which case the branch falls back to the majority of the whole trajectory. 
The absorbing chain of Appendix~\ref{app:dp} computes (\textit{a}) and (\textit{b}) exactly, as $U - W$ and $V$, for any candidate distribution $q$; evaluating it at $q^{\mathrm{pess}}$ and invoking Lemma~\ref{lem:dom} transfers the bound to every $p$ in the confidence set on $\Gcal$. 
For (\textit{c}), the trajectory majority at budget exhaustion is the majority over $E^*$ iid samples; the Hoeffding bound of Section~\ref{sec:theory-naive} with $\bar K$ in place of $K$ (Assumption~\ref{asm:color-bound}) and $d_{\mathrm{use}} \le \Delta$ on $\Gcal$ bounds its probability by $(\bar K - 1) e^{-N d_{\mathrm{use}}^2 / 2} = F$, which the chooser includes as its third term. 
Summing the three parts bounds $\Prob(\text{wrong} \mid \text{enter the fast branch}, \Gcal)$ by $p_{\mathrm{err}}[w; q^{\mathrm{pess}}] + \rho_N$, which the criterion of Theorem~\ref{thm:fast} charges to the $p^*$ budget. 

\subsection{Proof of the majority Hoeffding bound (Section~\ref{sec:theory-naive})}\label{app:proof-mode}

Fix $c \ge 2$ and define $Y_i = \mathbf 1\{X_i = c\} - \mathbf 1\{X_i = 1\}$; then $\E Y_i = -\Delta_c \le -\Delta$ and $|Y_i| \le 1$. 
The majority is wrong if and only if there exists $c \ge 2$ with $N_c \ge N_1$, that is $\sum_i Y_i \ge 0$, and Hoeffding's inequality on $[-1,1]$ gives $\Prob(\sum_i Y_i \ge 0) \le e^{-N\Delta_c^2/2} \le e^{-N\Delta^2/2}$. 
A union over $c = 2, \dots, K$ gives at most $(K-1)e^{-N\Delta^2/2}$, and replacing $K$ by its upper bound $\bar K$ gives the bound used in the text. 

\subsection{Proof of the answering-error and mis-refusal bounds (Theorem~\ref{thm:error})}\label{app:proof-error}

Define the good event $\Gcal$ with $\Prob(\Gcal) \ge 1-\delta$ (Lemma~\ref{lem:cs} and Corollary~\ref{cor:trigger}); on $\Gcal$, $|\hat\Delta_t - \Delta| \le w_t$ for all $t \ge 2$, hence $d_{\mathrm{use}} = \max(0, \hat\Delta_\tau - w_\tau) \le \Delta$. 

\textbf{Answering error (joint).} The rule answers through the fast branch or through the naive branch. 
On $\Gcal$ both paths are safe: Theorem~\ref{thm:fast} bounds the fast-path error by $p^*$, and in the naive branch the criterion $(\bar K-1)e^{-N d_{\mathrm{use}}^2/2} \le p^*$ together with $d_{\mathrm{use}} \le \Delta$ and $K \le \bar K$ implies $(K-1)e^{-N\Delta^2/2} \le p^*$, so the true error is at most $p^*$ as well. 
Hence $\Prob(\text{wrong} \wedge \text{answers} \wedge \Gcal) \le p^* \Prob(\text{answers} \wedge \Gcal)$, and off $\Gcal$ only the mass $\delta$ remains, so
\[
\Prob(\text{wrong} \wedge \text{answers}) \;\le\; p^* \Prob(\text{answers} \wedge \Gcal) + \delta \;\le\; p^* + \delta .
\]

\textbf{Mis-refusal.} A mis-refusal is the joint event $\text{refuse} \wedge \mathcal{A}$, with $\mathcal{A} = \{(K-1)e^{-N\Delta^2/2} \le p^*\}$ the naive-feasibility event of Section~\ref{sec:method-formulation}. 
Any mis-refusal either lies outside $\Gcal$ --- probability at most $\delta$ --- or lies inside $\Gcal$, where by definition it is counted by the plug-in bias $B_{\mathrm{plug}}$: on $\Gcal$, a refusal means $(\bar K-1)e^{-N d_{\mathrm{use}}^2/2} > p^*$, which is compatible with $\mathcal{A}$ only if the conservative endpoint $d_{\mathrm{use}}$ underestimates $\Delta$ enough to flip the criterion, or if the replacement of $K$ by $\bar K$ does so. 
Hence $\Prob(\text{refuse} \wedge \mathcal{A}) \le \delta + B_{\mathrm{plug}}$. 
The sufficient condition for $B_{\mathrm{plug}} = 0$ is therefore $d_{\mathrm{use}} = \Delta$ \emph{and} $K = \bar K$; the color-count bound contributes a mis-refusal channel of its own when $K < \bar K$, which no distribution-free argument can remove and which we therefore report rather than bound. 

\subsection{Proof of the expected-samples decomposition (Theorem~\ref{thm:expected})}\label{app:proof-expected}

The three terms correspond to the three branches. 
The fast branch spends $\tau_{\mathrm F}$ in exploration and then an expected $\min(T_{\mathrm{run}}, N)$ in the run phase, truncated by the remaining budget; the naive branch always spends $E^*$ in total; the refusal branch spends $\tau_{\mathrm R}$. 
Weighting by the branch probabilities and applying linearity of expectation gives the decomposition. 
Because the majority's run typically completes first in the multi-color race, the true expectation is slightly below $\E[T_{\mathrm{run}}(\hat p_{(1)}, w)]$, so the stated fast-branch term is an upper bound. 

\subsection{Proof of the two-point lower bound (Theorem~\ref{thm:infolb})}\label{app:proof-infolb}

We use the Neyman--Pearson log-likelihood-ratio test. 
Let $\Lambda_n = \sum_{i=1}^{n} \ln\frac{dP^{+}}{dP^{-}}(X_i)$, a random walk with mean $n\,\mathrm{KL}$ under $P^{+}$ and $-n\,\mathrm{KL}$ under $P^{-}$; decide $P^{+}$ if and only if $\Lambda_n \ge 0$. 
By Stein's lemma the sum of the two error probabilities of the optimal test decays as $e^{-n\,\mathrm{KL}(1+o(1))}$, so requiring per-side error at most $\delta$ gives $n\,\mathrm{KL} \ge \ln\frac{1}{2\delta}(1+o(1))$. 
Substituting the two-point family's divergence $\mathrm{KL} = \Delta\ln\frac{1+\Delta}{1-\Delta} = 2\Delta^2 + O(\Delta^4)$ yields $n \ge c\Delta^{-2}\ln\frac{1}{\delta}$ for an absolute constant $c$. 
The sequential version follows by applying the same argument at every time step, a standard technique in bandit lower bounds; the expected stopping time has the same order. 

\section{The Choose-$w$ dynamic program, and the oracle comparison}\label{app:dp}

\subsection{The chain behind $p_{\mathrm{err}}$}\label{app:dp-chain}

The run part is computed by an absorbing Markov chain over states $(c, r)$: the last $r$ draws are all of color $c$ and no $w$-run has been completed. 
Let $g(c, r)$ be the probability that, starting from $(c, r)$, the first completed $w$-run is completed by a non-majority color, with absorption $g(c, w) = \mathbf 1\{c \text{ is not the majority}\}$. 
Then for $r = 1, \dots, w-1$,
\[
\begin{aligned}
g(c, r) &= q_c\, g(c, r+1) + \sum_{c' \neq c} q_{c'}\, g(c', 1),\\
g(\emptyset) &= \sum_{c} q_c\, g(c, 1).
\end{aligned}
\]
This system has $O(\hat K_\tau w)$ states and a rank-one coupling, and it can be solved in $O(w\hat K_\tau)$ time by exploiting its near-triangular structure. 
The count-gap terms $V$ and $W$ require, in addition, the count gap $d = \hat N_{(1)} - \max_{c \neq (1)} \hat N_c$ between the majority color and the strongest minority color; the augmented chain over the run configuration and $d$, with $|d| \le N$, has $O(\hat K_\tau wN)$ states in the worst case, which for the folded two-color distributions used by the pessimistic mode reduces to $O(wN)$. 
The fallback term is bounded by the majority Hoeffding bound of Section~\ref{sec:theory-naive}, which the chooser adds as its third term. 
One modelling remark is in order, because it delimits what the dynamic program proves: the count gap $d$ is treated as a sufficient state together with the run configuration, which is exact for the two-color folds the pessimistic mode evaluates and is an approximation for a general multi-color empirical distribution. 
All of this runs on the CPU and is negligible relative to the sampling cost (Section~\ref{sec:method-complexity}). 

\subsection{Comparison with an omniscient oracle}\label{app:oracle}

We record the oracle comparison here because it is a supporting computation rather than part of the guarantee chain. 
Let $C_{\mathrm{adapt}}$ and $C_{\mathrm{oracle}}$ denote the expected sample counts of the rule and of an oracle that knows $p$ and therefore skips exploration, and let $w^*$ and $w^{\mathrm{pess}}$ be the smallest windows satisfying $p_{\mathrm{err}}[w^*; p] \le p^*$ and $p_{\mathrm{err}}[w; q^{\mathrm{pess}}] + \rho_N \le p^*$ respectively. 

\begin{theorem}[Oracle gap, achievable instances]\label{thm:oracle}
For instances on which both rules take the fast branch,
\[
\begin{aligned}
0 \;\le\; C_{\mathrm{adapt}} - C_{\mathrm{oracle}} \;\le\;& T_{\mathrm{cap}} + \E[T_{\mathrm{run}}(\hat p_{(1)}, w^{\mathrm{pess}})]\\
& - \E[T_{\mathrm{run}}(p_{(1)}, w^*)] + O(\delta E^*).
\end{aligned}
\]
\end{theorem}

The three components are the exploration cap, which the oracle does not pay, the additional waiting time induced by the conservative window, and the expected cost of mis-triggers. 
The proof proceeds in three steps: the pessimistic feasible set is contained in the true one, so $w^{\mathrm{pess}} \ge w^*$; on achievable instances the two rules differ term by term along the branch decomposition; and outside $\Gcal$, of probability at most $\delta$, each path costs at most $E^*$. 

\textbf{Step 1 ($w^{\mathrm{pess}} \ge w^*$).} On $\Gcal$, Lemma~\ref{lem:dom} gives $p_{\mathrm{err}}[w; p] \le p_{\mathrm{err}}[w; q^{\mathrm{pess}}] + \rho_N$ for every $w$ and every $p$ in the confidence set, so any window accepted by the pessimistic criterion is feasible for $p$; the pessimistic feasible set is therefore a subset of the true one. 

\textbf{Step 2 (branch-conditional comparison).} Conditional on branch entry at time $\tau$ with remaining budget $N = E^* - \tau$, Theorem~\ref{thm:expected} and the run-waiting formula give
\[
\begin{aligned}
C_{\mathrm{adapt}} - C_{\mathrm{oracle}} \;\le\;& T_{\mathrm{cap}} + \E[T_{\mathrm{run}}(\hat p_{(1)}, w^{\mathrm{pess}})]\\
& - \E[T_{\mathrm{run}}(p_{(1)}, w^*)].
\end{aligned}
\]
where $T_{\mathrm{cap}}$ bounds the exploration cost, the truncation of the waiting time by $N$ only lowers the adaptive term, and the expectations use the plug-in majority probability at entry. 

\textbf{Step 3 (mis-triggers).} Off $\Gcal$ the plug-in estimates may be uninformative and each such path spends at most $E^*$, contributing $O(\delta E^*)$. 

\section{Competing rules: exact stopping criteria}\label{app:competitors}

\begin{table*}[htbp]
\caption{Behavior across the difficulty sweep, 2000 seeds per cell, all rules on the same instances. ``Answer'' is the answer rate, ``cond.\ err.'' the error on answered instances (for the rules with no refusal output this equals the total error), ``miss (feas.)'' the rate at which a feasible instance is refused, and ``OK (infeas.)'' the rate at which an infeasible instance is correctly refused; the last two are defined only for BARS.}
\label{tab:delta-sweep}
\footnotesize
\setlength{\tabcolsep}{4pt}
\centering
\begin{tabular}{@{}lccccccc@{}}
\toprule
$\Delta$ & Rule & Answer & Cond.\ err. & Total err. & Mean $n$ & Miss (feas.) & OK (infeas.) \\
\midrule
0.02 & BARS (practical) & 0.134 & 0.3806 & 0.9170 & 134.0 & 0.0000 & 0.8660 \\
0.02 & Fixed-N & 1.000 & 0.3345 & 0.3345 & 200.0 & --- & --- \\
0.02 & SR & 1.000 & 0.4445 & 0.4445 & 200.0 & --- & --- \\
0.02 & SH & 1.000 & 0.4160 & 0.4160 & 150.0 & --- & --- \\
0.02 & LUCB & 1.000 & 0.3760 & 0.3760 & 200.0 & --- & --- \\
0.02 & lil'UCB & 1.000 & 0.3850 & 0.3850 & 200.0 & --- & --- \\
\midrule
0.05 & BARS (practical) & 0.161 & 0.2617 & 0.8815 & 131.1 & 0.0000 & 0.8395 \\
0.05 & Fixed-N & 1.000 & 0.2110 & 0.2110 & 200.0 & --- & --- \\
0.05 & SR & 1.000 & 0.3720 & 0.3720 & 200.0 & --- & --- \\
0.05 & SH & 1.000 & 0.3490 & 0.3490 & 150.0 & --- & --- \\
0.05 & LUCB & 1.000 & 0.2185 & 0.2185 & 200.0 & --- & --- \\
0.05 & lil'UCB & 1.000 & 0.3215 & 0.3215 & 200.0 & --- & --- \\
\midrule
0.1 & BARS (practical) & 0.240 & 0.1167 & 0.7880 & 124.4 & 0.0000 & 0.7600 \\
0.1 & Fixed-N & 1.000 & 0.0570 & 0.0570 & 200.0 & --- & --- \\
0.1 & SR & 1.000 & 0.2490 & 0.2490 & 200.0 & --- & --- \\
0.1 & SH & 1.000 & 0.1580 & 0.1580 & 150.0 & --- & --- \\
0.1 & LUCB & 1.000 & 0.0630 & 0.0630 & 200.0 & --- & --- \\
0.1 & lil'UCB & 1.000 & 0.1975 & 0.1975 & 200.0 & --- & --- \\
\midrule
0.15 & BARS (practical) & 0.316 & 0.0380 & 0.6960 & 117.8 & 0.0000 & 0.6840 \\
0.15 & Fixed-N & 1.000 & 0.0115 & 0.0115 & 200.0 & --- & --- \\
0.15 & SR & 1.000 & 0.1865 & 0.1865 & 200.0 & --- & --- \\
0.15 & SH & 1.000 & 0.0830 & 0.0830 & 150.0 & --- & --- \\
0.15 & LUCB & 1.000 & 0.0130 & 0.0130 & 200.0 & --- & --- \\
0.15 & lil'UCB & 1.000 & 0.1330 & 0.1330 & 200.0 & --- & --- \\
\midrule
0.2 & BARS (practical) & 0.472 & 0.0244 & 0.5395 & 102.9 & 0.0000 & 0.5280 \\
0.2 & Fixed-N & 1.000 & 0.0000 & 0.0000 & 200.0 & --- & --- \\
0.2 & SR & 1.000 & 0.1370 & 0.1370 & 200.0 & --- & --- \\
0.2 & SH & 1.000 & 0.0360 & 0.0360 & 150.0 & --- & --- \\
0.2 & LUCB & 1.000 & 0.0025 & 0.0025 & 200.0 & --- & --- \\
0.2 & lil'UCB & 1.000 & 0.0730 & 0.0730 & 200.0 & --- & --- \\
\midrule
0.3 & BARS (practical) & 0.821 & 0.0043 & 0.1825 & 64.8 & 0.1790 & --- \\
0.3 & Fixed-N & 1.000 & 0.0000 & 0.0000 & 200.0 & --- & --- \\
0.3 & SR & 1.000 & 0.0995 & 0.0995 & 200.0 & --- & --- \\
0.3 & SH & 1.000 & 0.0040 & 0.0040 & 150.0 & --- & --- \\
0.3 & LUCB & 1.000 & 0.0000 & 0.0000 & 200.0 & --- & --- \\
0.3 & lil'UCB & 1.000 & 0.0150 & 0.0150 & 200.0 & --- & --- \\
\midrule
0.4 & BARS (practical) & 0.973 & 0.0005 & 0.0275 & 38.6 & 0.0270 & --- \\
0.4 & Fixed-N & 1.000 & 0.0000 & 0.0000 & 200.0 & --- & --- \\
0.4 & SR & 1.000 & 0.0600 & 0.0600 & 200.0 & --- & --- \\
0.4 & SH & 1.000 & 0.0000 & 0.0000 & 150.0 & --- & --- \\
0.4 & LUCB & 1.000 & 0.0000 & 0.0000 & 199.8 & --- & --- \\
0.4 & lil'UCB & 1.000 & 0.0015 & 0.0015 & 200.0 & --- & --- \\
\midrule
0.6 & BARS (practical) & 1.000 & 0.0000 & 0.0000 & 18.5 & 0.0000 & --- \\
0.6 & Fixed-N & 1.000 & 0.0000 & 0.0000 & 200.0 & --- & --- \\
0.6 & SR & 1.000 & 0.0260 & 0.0260 & 200.0 & --- & --- \\
0.6 & SH & 1.000 & 0.0000 & 0.0000 & 150.0 & --- & --- \\
0.6 & LUCB & 1.000 & 0.0000 & 0.0000 & 166.9 & --- & --- \\
0.6 & lil'UCB & 1.000 & 0.0000 & 0.0000 & 200.0 & --- & --- \\
\midrule
0.8 & BARS (practical) & 1.000 & 0.0000 & 0.0000 & 14.0 & 0.0000 & --- \\
0.8 & Fixed-N & 1.000 & 0.0000 & 0.0000 & 200.0 & --- & --- \\
0.8 & SR & 1.000 & 0.0050 & 0.0050 & 200.0 & --- & --- \\
0.8 & SH & 1.000 & 0.0000 & 0.0000 & 150.0 & --- & --- \\
0.8 & LUCB & 1.000 & 0.0000 & 0.0000 & 96.7 & --- & --- \\
0.8 & lil'UCB & 1.000 & 0.0000 & 0.0000 & 200.0 & --- & --- \\
\bottomrule
\end{tabular}
\end{table*}

Every competitor is implemented from its published stopping rule and consumes the same sampled stream as BARS: one sample is one draw from the underlying distribution (synthetic setting) or one parsed decode (LLM setting). 
For a stream $x_1, x_2, \dots$ we write $N_c(t)$ for the count of value $c$ up to time $t$, $\hat p_c(t) = N_c(t)/t$, and $\hat p_{(1)}(t) \ge \hat p_{(2)}(t)$ for the top two empirical frequencies. 

\subsection{Fixed-budget and fixed-confidence best-arm identification}

\textbf{\textsc{Fixed-$N$}.} 
Stop at $t = N$ and return $\arg\max_c N_c(N)$; the rule never refuses and always spends the whole budget. 

\textbf{Successive Rejects \citep{kaufmann2016}.} 
With budget $B$ and $K$ arms, set $\bar n = K + \tfrac12 - \sum_{i=2}^{K} \tfrac1i$, and in round $k = 1,\dots,K-1$ draw $n_k = \lceil (B-K)/(\bar n\,(K+1-k)) \rceil$ times from each surviving arm, then eliminate the arm with the smallest empirical mean. 

\textbf{Sequential Halving \citep{kalyanakrishnan2012}.} 
Split the budget uniformly over $\lceil \log_2 K\rceil$ rounds and, in each round, draw equally from the survivors and keep the better half by empirical mean. 

\textbf{lil'UCB \citep{jamieson2016}.} 
Round-robin sampling with the radius
$\mathrm{rad}_a(t) = \sqrt{1.5\ln\!\big((\ln n_a)^2 K\, 4t^2/\delta\big)/n_a}$;
stop when $\hat p_a - \mathrm{rad}_a > \max_{b \ne a}(\hat p_b + \mathrm{rad}_b)$ for the leader $a$. 

\textbf{LUCB \citep{kalyanakrishnan2012}.} 
In each round sample the empirical leader $h$ and runner-up $\ell$ once, and stop when
$\hat p_h - \mathrm{rad}_h > \hat p_\ell + \mathrm{rad}_\ell$ with the Hoeffding radius
$\mathrm{rad}_a = \sqrt{\ln(4Kt^2/\delta)/(2n_a)}$. 

\textbf{Hoeffding race \citep{maron1993}.} 
In each round draw once from every surviving arm and eliminate any arm $a$ with
$\hat p_a + \mathrm{rad}_a < \hat p_{\mathrm{best}} - \mathrm{rad}_{\mathrm{best}}$, where
$\mathrm{rad}_a = \sqrt{\ln(2K/\delta)/(2n_a)}$. 

All five rules are run under the same per-path budget $E^*$ and return their leader at expiry; \textsc{Fixed-$N$} and the elimination rules have no refusal output, which is why Section~\ref{sec:experiments-design} reports their conditional and total error as one number. 

\subsection{Sequential early-stopping rules for LLM majority voting}

\textbf{ESC \citep{esc2024}.} 
Given a window $w$ and a cap $L$, stop at the first $t \ge w$ with
$x_{t-w+1} = \dots = x_t$ (a window in which all answers coincide), and answer with the plurality over the prefix; if no such window occurs, answer with the plurality over $L$ samples. 
The paper's control scheme fixes $w$ from a warmup window of size $w_0 = 5$; we report $w \in \{2,3,4\}$, which brackets the values its figures use. 

\textbf{CGES \citep{cges2025}.} 
The posterior score of candidate $a_i$ after $m$ samples is
$s_{a_i} = \prod_{t:\,R_t = a_i} C_t \prod_{t:\,R_t \ne a_i} \frac{1-C_t}{K-1}$, normalised over candidates; the rule stops at the first $t$ with $\max_i s_{a_i} \ge \gamma$. 
The score depends on a per-sample decoding confidence $C_t$, which the pool-replay protocol does not carry, so we evaluate the confidence-free reduction
$s_{a_i} \propto N_{a_i}(t) + 1$ (the Dirichlet-multinomial posterior mean under a uniform prior, with an explicit unseen-answer bucket so that $K \ge 2$ from the first round, as the paper's online implementation does) and stop on $\max_i \hat p_{a_i}(t) \ge \gamma$ with $\gamma \in \{0.9, 0.95\}$. 
This reduction is stated in Section~\ref{sec:experiments-design} and is the reason the CGES numbers here are a lower bound on its advantage rather than an evaluation of its confidence weighting. 

\textbf{DASE \citep{dase2026}.} 
DASE commits when the evidence is concentrated and otherwise falls back to the global frequency; on a pool without deliberation rounds we instantiate it as committing at the first $t$ with $\hat p_{(1)}(t) \ge 1/2 + \tau$ and $\tau = 0.1$, and falling back to the plurality at $L$. 

\textbf{MARS \citep{mars2026}.} 
The margin between the leader and the runner-up must remain safe under a conservative bound on future vote movement; at the answer level we use
$\mathrm{margin}(t) = N_{(1)}(t) - N_{(2)}(t) > z\sqrt{2\hat\sigma^2 t}$ with $z = 2$ and $\hat\sigma^2$ the plug-in variance of Section~\ref{sec:method-notation}, stopping at the first $t$ where this holds. 
MARS's real rule also learns per-trace switch probabilities, which need partial-trace states that the pool does not carry. 

\textbf{MMC, the certificate of certified self-consistency \citep{certifiedsc2025}.} 
The Martingale Majority Certificate is an anytime-valid stopping rule built on $e$-processes; at the answer level it is the certified width of Section~\ref{sec:method-width} applied without the budget or refusal logic,
$w_t = \sqrt{2\hat\sigma_t^2\,(2\ln t + \ln\tfrac{\pi^2}{3\delta})/t}$, with a stop at the first $t$ where $\hat\Delta_t - w_t \ge \Delta_{\mathrm{high}}$. 
The distinction from BARS is not the width but the absence of a refusal output and of a budget argument: MMC answers at $L$ whatever the width says, which is why its conditional error is reported on the full pool. 

\end{document}